\documentclass[12pt]{iopart}

\usepackage{graphicx}
\usepackage{overpic}
\usepackage{multirow}
\usepackage{hyperref}

\usepackage{xcolor}
\newcommand{\column}[2]{%
	\begin{tabular}[b]{@{}c@{}}#1\\#2\end{tabular}%
}
\begin{document}

\title{Event-based Shape from Polarization with Spiking Neural Networks}

\author{Peng Kang\,$^{1,*}$, Srutarshi Banerjee$^{2}$, Henry Chopp$^{3}$, Aggelos Katsaggelos$^{3}$, and Oliver Cossairt$^{1}$}

\address{$^{1}$Department of Computer Science, Northwestern, Evanston, IL, USA \\
	$^{2}$Argonne National Laboratory, Lemont, IL, USA \\
	$^{3}$Department of Electrical and Computer Engineering, Northwestern, Evanston, IL, USA}
\ead{pengkang2022@u.northwestern.edu}
\vspace{10pt}
\begin{indented}
\item[]Dec. 2023
\end{indented}

\begin{abstract}
Recent advances in event-based shape determination from polarization offer a transformative approach that tackles the trade-off between speed and accuracy in capturing surface geometries. In this paper, we investigate event-based shape from polarization using Spiking Neural Networks (SNNs), introducing the Single-Timestep and Multi-Timestep Spiking UNets for effective and efficient surface normal estimation. Specificially, the Single-Timestep model processes event-based shape as a non-temporal task, updating the membrane potential of each spiking neuron only once, thereby reducing computational and energy demands. In contrast, the Multi-Timestep model exploits temporal dynamics for enhanced data extraction. Extensive evaluations on synthetic and real-world datasets demonstrate that our models match the performance of state-of-the-art Artifical Neural Networks (ANNs) in estimating surface normals, with the added advantage of superior energy efficiency. Our work not only contributes to the advancement of SNNs in event-based sensing but also sets the stage for future explorations in optimizing SNN architectures, integrating multi-modal data, and scaling for applications on neuromorphic hardware.
\end{abstract}

%
%
%
%
%
\section{Introduction}
Precise surface normal estimation can provide valuable information about a scene’s geometry and is useful for many computer vision tasks, including 3D Reconstruction~\cite{kazhdan2006poisson}, Augmented Reality (AR) and Virtual Reality (VR)~\cite{luo2020niid, ju2021recovering}, Material Classification~\cite{strese2016multimodal}, and Robotics Navigation~\cite{badino2011fast}. Depending upon the requirements of the application, surface normal estimation can be carried out using a variety of methods~\cite{geng2011structured, horn1970shape, shafer1983using, witkin1981recovering, hernandez2004stereo, woodham1980photometric}. In this work, we are interested in estimating surface normal from polarization images -- shape from polarization~\cite{rahmann2001reconstruction, kadambi2015polarized, wolff1993constraining, tozza2017linear, ba2020deep, lei2022shape}. In particular, shape from polarization leverages the polarization state of light to infer the shape of objects. When light reflects off surfaces, it becomes partially polarized. This method uses this property to estimate the surface normals of objects, which are then used to reconstruct their 3D shape. Compared to other 3D sensing methods, shape from polarization has many advantages~\cite{kadambi2015polarized, kadambi2017depth}, such as its suitability for capturing fine details on a variety of surface materials, including reflective and transparent ones, and its reliance on passive sensing, which eliminates the need for external light sources or emitters. Additionally, shape from polarization can provide high-precision data with relatively low-cost and low-energy equipment, making it an efficient and versatile option for 3D imaging in various applications.

Typically, a polarizing filter is used in conjunction with a camera to capture the polarization images and infer the polarization information. Generally, there are two ways to capture the polarization images and estimate the surface normals from them, one is \textit{Division of Time (DoT)}~\cite{kadambi2015polarized, atkinson2018high, wolff1997polarization} and the other one is \textit{Division of Focal Plane (DoFP)}~\cite{ba2020deep, lei2022shape, LucidV}. The DoT approaches add a rotatable linear polarizer in front of the lens of an ordinary camera. The filter is rotated to different orientations, and full-resolution polarization images are captured for each orientation at different times. By analyzing the changes in the polarization state of light across these images, the surface normals of objects can be estimated. The DoT methods use the full resolution of the sensor but trade-off against acquisition time. On the other hand, the DoFP methods place an array of micro-polarizers in front of the camera~\cite{LucidV}. This allows the camera to capture polarization information at different orientations in a single shot. Despite the reduced latency, this system is limited by the low resolution of polarization images, as each pixel only captures polarization at a specific orientation. This can result in lower accuracy compared to the DoT methods.

To bridge the accuracy of DoT with the speed of DoFP, researchers propose event-based shape from polarization following the DoT design scheme~\cite{muglikar2023event}. Specifically, a polarizer is rotating in front of an event camera~\cite{gallego2020event} and this creates sinosoidal changes in brightness intensity. Unlike traditional DoT methods utilize standard cameras to capture full-resolution polarization images at fixed rates, event-based shape from polarization employs event cameras to asynchronously measure changes in brightness intensity for each pixel within the full-resolution scene and trigger the events with microsecond resolution if the difference in brightness exceeds a threshold. The proposed event-based method uses the continuous event stream to reconstruct relative intensities at multiple polarizer angles. These reconstructed polarized images are then utilized to estimate surface normals using physics-based and learning-based methods~\cite{muglikar2023event}. Due to the DoT-driven characteristic and low latency event cameras provide, the event-based shape from polarization mitigates the accuracy-speed trade-off in the traditional shape from polarization field.

Although the event-based shape from polarization brings many advantages, we still need to carefully choose models that process the data from event cameras. With the prevalence of Artifical Neural Networks (ANNs), one recent method~\cite{muglikar2023event} employs ANNs to process event data and demonstrates the better surface normal estimation performance compared to physics-based methods. However, ANNs are not compatiable with the working mechanism of event cameras and incur the high energy consumption. To be more compatiable with event cameras and maintain the high energy efficiency, research on Spiking Neural Networks (SNNs)~\cite{roy2019towards} starts to gain momentum. Similar to event cameras that mimic the human retina's way of responding to changes in light intensity, SNNs are also bio-inspired and designed to emulate the neural dynamics of human brains. Unlike ANNs employing artificial neurons~\cite{maas2013rectifier, xu2015empirical, clevert2015fast} and conducting real-valued computation, SNNs adopt spiking neurons~\cite{gerstner1995time, abbott1999lapicque, gerstner2002spiking} and utilize binary 0-1 spikes to process information. This difference reduces the mathematical dot-product operations in ANNs to less computationally summation operations in SNNs~\cite{roy2019towards}. Due to such the advantage, SNNs are always energy-efficient and suitable for power-constrained devices. Although SNNs demonstrate the higher energy efficiency and much dedication has been devoted to SNN research, ANNs still present the better performance and dominate in a wide range of learning applications~\cite{lecun2015deep}. 

Recently, more research efforts have been invested to shrink the performance gap between ANNs and SNNs. And SNNs have achieved comparable performance in various tasks, including image classification~\cite{zhou2022spikformer}, object detection~\cite{zhang2022spiking}, graph prediction~\cite{zhu2022spiking}, natural language processing~\cite{zhu2023spikegpt}, etc. Nevertheless, we have not yet witnessed the establishment of SNN in the accurate surface normal estimation with an advanced performance. To this end, this naturally raises an issue: \textit{could bio-inspired Spiking Neural Networks estimate surface normals from event-based polarization data with an advanced quality at low energy consumption?} 

In this paper, we investigate the event-based shape from polarization with a spiking approach to answer the above question. Specifically, inspired by the feed-forward UNet~\cite{ronneberger2015u} for event-based shape from polarization~\cite{muglikar2023event}, we propose the Single-Timestep Spiking UNet, which treats the event-based shape from polarization as a non-temporal task. This model processes event-based inputs in a feed-forward manner, where each spiking neuron in the model updates its membrane potential only once. Although this approach may not maximize the temporal processing capabilities of SNNs, it significantly reduces the computational and energy requirements. To further exploit the rich temporal information from event-based data and enhance model performance in the task of event-based shape from polarization, we propose the Multi-Timestep Spiking UNet. This model processes inputs in a sequential, timestep-by-timestep fashion, allowing each spiking neuron to utilize its temporal recurrent neuronal dynamics to more effectively extract information from event data. We extensively evaluate the proposed models on the synthetic dataset and the real-world dataset for event-based shape from polarization. The results of these experiments, both quantitatively and qualitatively, indicate that our models are capable of estimating dense surface normals from polarization events with performance comparable to current state-of-the-art ANN models. Additionally, we perform ablation studies to assess the impact of various design components within our models, further validating their effectiveness. Furthermore, our models exhibit superior energy efficiency compared to their ANN counterparts, which highlights their potential for application on neuromorphic hardware and energy-constrained edge devices. 

The remainder of this paper is structured as follows: Section II provides a comprehensive review of existing literature on shape from polarization and SNNs. In Section III, we detail our proposed SNN models for event-based shape from polarization, including their structures, training protocols, and implementation details. Section IV showcases the effectiveness and energy efficiency of our proposed models on different benchmark datasets. The paper concludes with Section V, where we summarize our findings and outline potential avenues for future research. 


%

\section{Related Work}
In the following, we will first give an overview of the related work on shape from polarization, including the traditional shape from polarization and event-based shape from polarization. Then, we will give a comprehensive review of SNNs and their applications in 3D scenes. 

\subsection{Shape from Polarization}
Shurcliff proposed the method of shape recovery by polarization information in 1962~\cite{shurcliff1962polarized}. Essentially, when unpolarized light reflects off a surface point, it becomes partially polarized. And the observed scene radiance varies with changing the polarizer angle, which encodes some relationship with surface normals. Therefore, by analyzing such relationship at each surface point through Fresnel equations~\cite{collett2005field}, shape from polarization methods can measure the azimuthal and zenithal angles at each pixel and recover the per-pixel surface normal with high resolution. Generally, two schemes are utilized to collect polarization images. One is \textit{Division of Time (DoT)}~\cite{kadambi2015polarized, atkinson2018high, wolff1997polarization} that provides full-resolution polarization images but increases the acquisition time significantly, while the other one is \textit{Division of Focal Plane (DoFP)}~\cite{ba2020deep, lei2022shape, LucidV} that trade-offs spatial resolution for low latency. After collecting the polarization images, various physical-based or learning-based methods~\cite{shi2020recent} can be utilized to estimate the surface normals. However, since a linear polarizer cannot distinguish between polarized light that is rotated by $\pi$ radians, this results in two confounding estimates for azimuth angle at each pixel~\cite{ba2020deep, miyazaki2003polarization}. To solve such the ambiguity, we have to carefully design the estimation methods by exploring additional constraints from various aspects, such as geometric cues~\cite{miyazaki2017surface, yang2018polarimetric, zhu2019depth}, spectral cues~\cite{wolff1993constraining, stolz2012shape, wolff1998improved}, photometric cues~\cite{tozza2017linear, yu2017shape, smith2018height}, or priors learned from deep learning techniques~\cite{ba2020deep, lei2022shape}.

Recently, with the prevalence of bio-inspired neuromorphic engineering, researchers begin to shift their focus to high-speed energy-efficient event cameras and propose solutions that combine polarization information with event cameras. Specifically, inspired by the polarization vision in the mantis shrimp eye~\cite{marshall1988unique}, \cite{haessig2023pdavis} proposed the PDAVIS polarization event camera. The researchers employed the DoFP scheme to design such the camera. This involved fabricating an array of pixelated polarization filters and strategically positioning them atop the sensor of an event camera. While this camera is adept at capturing high dynamic range polarization scenes with high speeds, it still faces challenges with low spatial resolution, a common issue inherent in the DoFP methods. To bridge the high resolution of DoT with the low latency of DoFP, \cite{muglikar2023event} adopted the DoT scheme and collected polarization events by placing a rotating polarizing filter in front of an event camera. Due to the high resolution of DoT and the low latency of event cameras, this method facilitates shape from polarization at both high speeds and with high spatial resolution. Typically, the captured polarization events are transformed into frame-like event representations~\cite{zhu2019unsupervised}, which are then processed using ANN models~\cite{muglikar2023event} to estimate surface normals. While these learning-based methods demonstrate superior performance over traditional physics-based methods, they significantly increase the energy consumption of the overall system, primarily due to the lower energy efficiency of ANNs. Through processing event polarization data collected by the promising DoT scheme, this paper aims to address this challenge by conducting event-based shape from polarization using SNNs, presenting a more energy-efficient alternative in this domain.




\subsection{Spiking Neural Networks}
With the development of ANNs, artificial intelligene models today have demonstrated extraordinary abilities in many tasks, such as computer vision, natural language processing, and robotics. Nevertheless, ANNs only mimic the brain's architecture in a few aspects, including vast connectivity and structural and functional organizational hierarchy~\cite{roy2019towards}. The brain has more information processing mechanisms like the neuronal and synaptic functionality \cite{bullmore2012economy, felleman1991distributed}. Moreover, ANNs are much more energy-consuming compared to human brains. To integrate more brain-like characteristics and make artificial intelligence models more energy-efficient, researchers propose SNNs, which can be executed on power-efficient neuromorphic processors like TrueNorth~\cite{merolla2014million} and Loihi~\cite{davies2021advancing}. Like ANNs, SNNs are capable of implementing common network architectures, such as convolutional and fully-connected layers, yet they distinguish themselves by utilizing spiking neuron models~\cite{gerstner2002spiking}, such as the Leaky Integrate-and-Fire (LIF) model~\cite{abbott1999lapicque} and the Spike Response Model (SRM)~\cite{gerstner1995time}. Due to the non-differentiability of these spiking neuron models, training SNNs can be challenging. However, progress has been made through innovative approaches such as converting pre-trained ANNs to SNNs~\cite{cao2015spiking,sengupta2019going} and developing methods to approximate the derivative of the spike function~\cite{wu2018spatio, ijcai2020-211}. Thanks to the developement of these optimization techniques, several models have been proposed recently to tackle the complex tasks in 3D scenes. Notably, StereoSpike~\cite{ranccon2022stereospike} and MSS-DepthNet~\cite{wu2022mss} have pioneered the development of deep SNNs for depth estimation, achieving performance on par with the state-of-the-art ANN models. Additionally, SpikingNeRF~\cite{yao2023spiking} has successfully adapted SNNs for radiance field reconstruction, yielding synthesis quality comparable to ANN baselines while maintaining high energy efficiency. In this paper, our emphasis is on employing SNNs to tackle event-based shape from polarization, aiming to establish a method that is not only effective but also more efficient for event-based surface normal estimation.

\section{Methods}

In this paper, we focus on building SNNs to estimate surface normals through the use of a polarizer paired with an event camera. In this setup, the polarizer is mounted in front of the event camera and rotates at a constant high speed driven by a motor. This rotation changes the illumination of the incoming light. Event cameras generate an asynchronous event $e_i = (x_i, y_i, t_i, p_i)$ when the illumination variation at a given pixel reaches a given contrast threshold $C$:
\begin{equation}
	L(x_i, y_i, t_i) - L(x_i, y_i, t_i - \Delta t_i) = p_iC,
\end{equation}
where $L\doteq log(I)$ is the log photocurrent ("brightness"), $p_i \in \{-1, +1\}$ is the sign of the brightness change, and $\Delta t_i$ is the time since the last event at the pixel $(x_i, y_i)$. The surface normal vector can be represented by its azimuth angle $\alpha$ and zenith angle $\theta$ in a spherical coordinate system. And the proposed models predict the surface normal $\mathbf{N}$ as a 3-channel tensor $\mathbf{N} = (\sin\theta\cos\alpha, \sin\theta\sin\alpha, \cos\theta)$ through the event steam.

\subsection{Input Event Representation}

\begin{figure}
	\includegraphics[width=\linewidth]{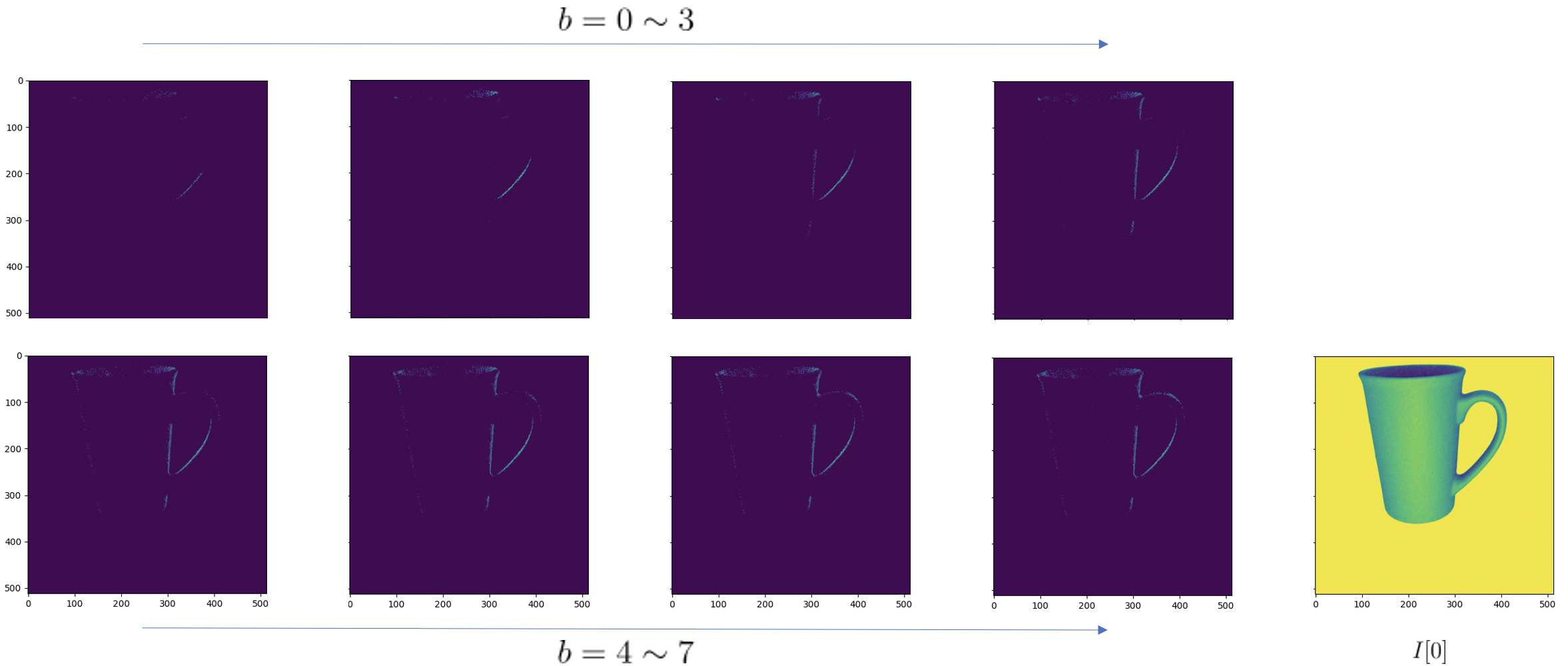}
	\caption{The CVGR-I input representation comprises CVGR frames spanning $B$ temporal bins, along with a single polarization image captured at a polarizer angle of 0 degrees. In this example, we set $B=8$.}
	\label{input}
\end{figure}

To ensure a fair comparison between our proposed methods and those utilizing ANNs for event-based shape from polarization, we transform the sparse event stream into frame-like event representations, which serve as the input for our methods. Specifically, similar to~\cite{muglikar2023event}, we take the CVGR-I representation due to its superior performance. The CVGR-I representation combines the Cumulative Voxel Grid Representation (CVGR) with a single polarization image (I) taken at a polarizer angle of 0 degrees. The CVGR is a variation of the voxel grid~\cite{zhu2019unsupervised}. Similar to previous works on learning with events~\cite{rebecq2019high, tulyakov2022time}, the CVGR first encodes the events in a spatio-temporal voxel grid $V$. Specifically, the time domain of the event stream is equally discretized into $B$ temporal bins indexed by integers in the range of $[0, B-1]$. Each event $e_i = (x_i, y_i, t_i, p_i)$ distributes its sign value $p_i$ to the two closest spatio-temporal voxels as follows:
\begin{equation}
	V(x, y, t) = \sum_{x_i = x, y_i = y} p_i \max(0, 1-|t-t_i^*|),\hspace{3mm}
	t_i^* = \frac{B-1}{\Delta T} (t_i - t_0),
\end{equation}
where $(x, y, t)$ is a specific location of the spatio-temporal voxel grid $V$, $\Delta T$ is the time domain of the event stream, and $t_0$ is the timestep of the initial event in the event stream. Then, the CVGR calculates the cumulative sum across the bins and multiplies this total by the contrast threshold:
\begin{equation}
	E(x, y, b) = C\sum_{i=0}^{b} V(x, y, i), \hspace{3mm}b = \{0, 1, 2, 3, ..., B-1\},
\end{equation}
Finally, to enhance surface normal estimation in areas with insufficient event information, a single polarization image of 0 polarizer degree is incorporated, resulting in $E = I[0] + E$, thereby providing additional context. This resulting event representation $E$ will serve as the input of our models. Its dimensions are $B\times H\times W$, where $H$ and $W$  represent the height and width of the event camera, respectively. We present a concrete input example of ``cup'' in Fig.~\ref{input}.


\subsection{Spiking Neuron Models}

Spiking neuron models are mathematical descriptions of specific cells in the nervous system. They are the basic building blocks of SNNs. In this paper, we primarily concentrate on using the Integrate-and-Fire (IF) model~\cite{abbott1999lapicque} to develop our proposed SNNs. The IF model is one of the earliest and simplest spiking neuron models. The dynamics of the IF neuron $i$ is defined as: 
\begin{equation}
	\label{eq:IF}
	u_i(t) = u_i(t-1) + \sum_jw_{ij}x_j(t),
\end{equation}
where $u_i(t)$ represents the internal membrane potential of the neuron $i$ at time $t$, $u_i(t-1)$ is the membrane potential of the neuron $i$ at the previous timestep $t-1$, and $\sum_jw_{ij}x_j(t)$ is the weighted summation of the inputs from pre-neurons at the current time step $t$. When $u_i(t)$ exceeds a certain threshold $u_{th}$, the neuron emits a spike, resets its membrane potential to $u_{reset}$, and then accumulates $u_i(t)$ again in subsequent time steps.

In addition to the IF model, we also build our proposed models with the Leaky Integrate-and-Fire (LIF) model~\cite{abbott1999lapicque}. Compared to the IF model, LIF model contains a leaky term to mimic the diffusion of ions through the membrane. The dynamics of the LIF neuron $i$ can be expressed as:
\begin{equation}
	u_i(t) = \alpha u_i(t-1) + \sum_jw_{ij}x_j(t),
\end{equation}
where $\alpha$ is a leaky factor that decays the membrane potential over time. Drawing inspiration from previous work~\cite{fang2021incorporating}, we also construct models using the Parametric Leaky Integrate-and-Fire (PLIF) model, which enables automatic learning of the leaky factor. In our experiments, we demonstrate that the IF model can offer better performance as it retains more information by not incorporating the leaky factor, thus striking a balance between high performance and biological plausibility.

\begin{figure}
	\includegraphics[width=\linewidth]{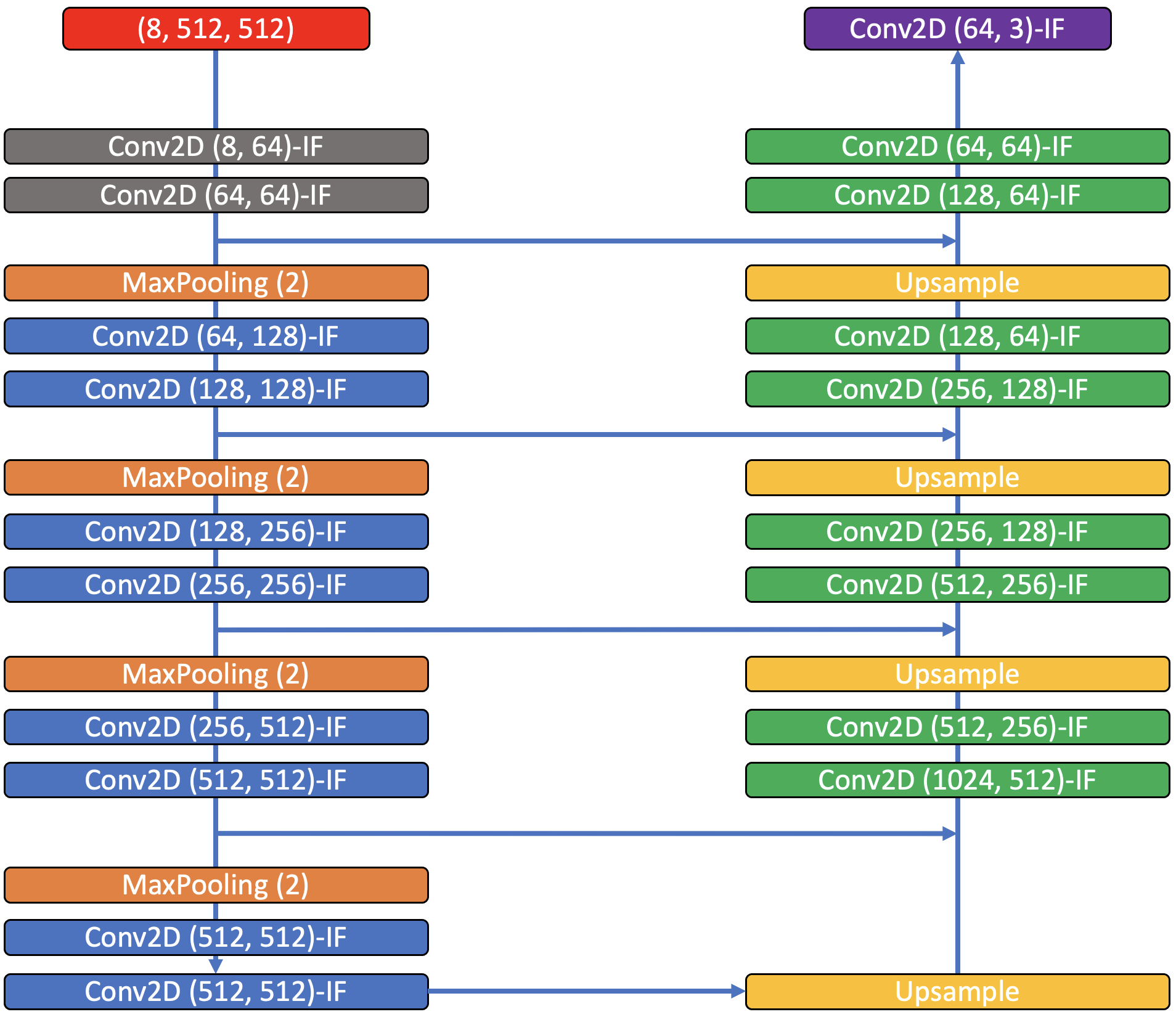}
	\caption{The network structure of Single-Timestep Spiking UNet: The network is designed according to the UNet architecture in a fully convolutional manner. Specifically, it consists of an event encoding module (gray), an encoder (orange and blue), a decoder (yellow and green), and a final prediction layer (purple). The size of CVGR-I input representation is $(8\times 512\times 512)$. Conv2D($a$, $b$)-IF represents the spiking convolutional layer with $a$ input channels and $b$ output channels. Each max pooling layer downsamples the feature map by a factor of 2. And the spatial resolution is doubled after each upsamling layer.}
	\label{single}
\end{figure}

\begin{figure}
	\includegraphics[width=\linewidth]{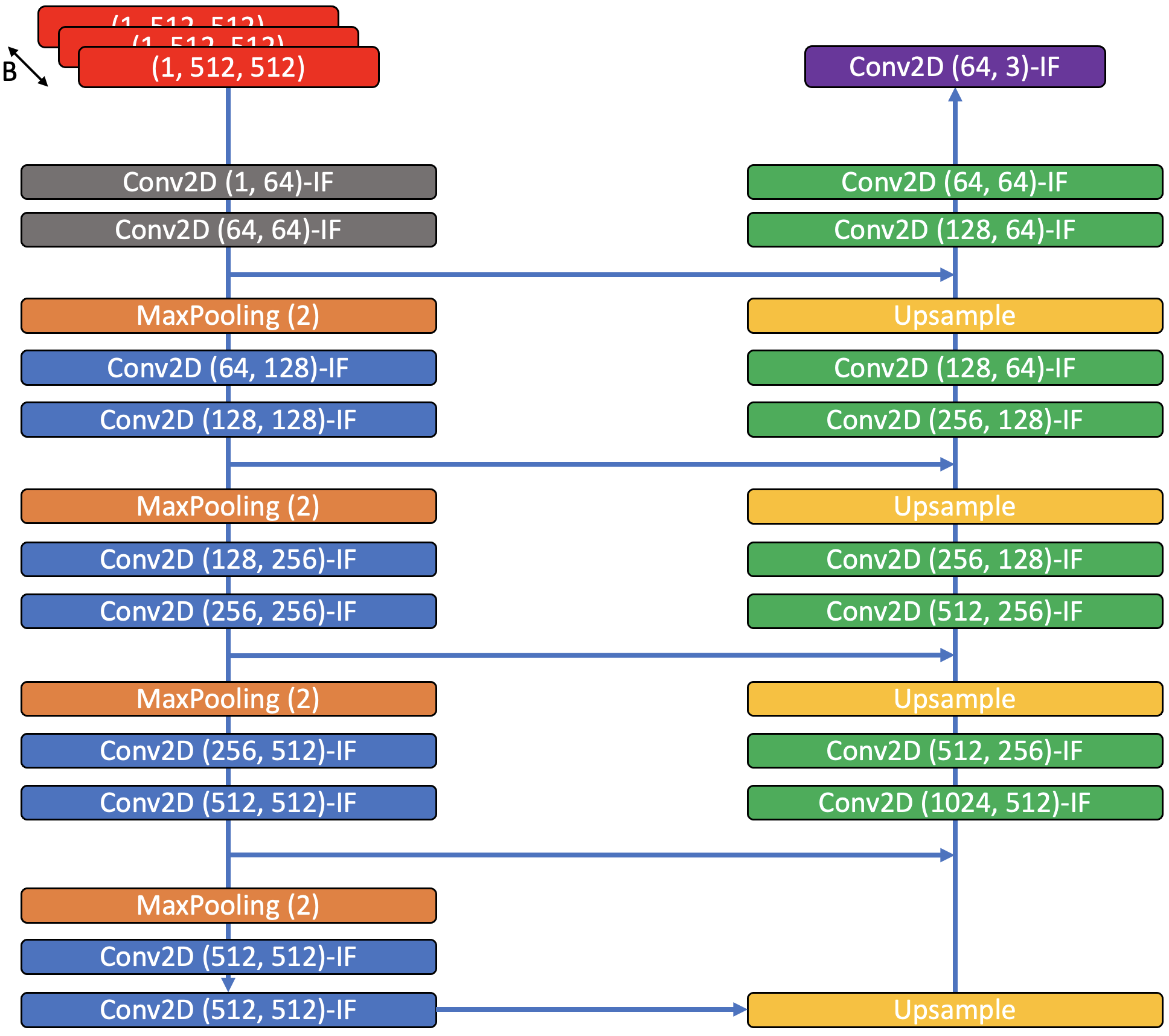}
	\caption{The network structure of Multi-Timestep Spiking UNet: The network is designed according to the UNet architecture in a fully convolutional manner. Specifically, it consists of an event encoding module (gray), an encoder (orange and blue), a decoder (yellow and green), and a final prediction layer (purple). Unlike the Single-Timestep Spiking UNet processing the CVGR-I representation as a whole and updating the membrane potential of its spiking neurons only once, the Multi-Timestep Spiking UNet processes the $B\times H\times W$ CVGR-I representation along its temporal dimension $B$. The settings for Conv2D($a$, $b$)-IF layers, max pooling layers, and upsampling layers are the same as those for the Single-Timestep Spiking UNet.}
	\label{multi}
\end{figure}

\subsection{SNNs for Event-based Shape from Polarization}

In this section, we propose two SNNs that take the CVGR-I event representation as the input and estimate the surface normals $\mathbf{N}$. Both of them can process the information through the spiking neuron models mentioned above. Due to the potential of IF neurons in event-based shape from polarization, we will present the proposed models based on the dynamics of IF neurons.

\subsubsection{Single-Timestep Spiking UNet}

In this work, we have chosen a UNet~\cite{ronneberger2015u}, a commonly utilized architecture in semantic segmentation, as the backbone for surface normal estimation. Specifically, we propose the Single-Timestep Spiking UNet as shown in Fig.~\ref{single}. This model is composed of several key components: an event encoding module, an encoder, a decoder, and a final layer dedicated to making surface normal predictions. As a Single-Timestep feed-forward SNN, this model processes the entire $B\times H\times W$ CVGR-I representation as its input and updates the membrane potential of its spiking neurons once per data sample. The event encoding module utilizes two spiking convolutional layers to transform the real-valued $B\times H\times W$ CVGR-I representation to the binary spiking representation with the size of $N_c\times H\times W$. Based on Eq.~\ref{eq:IF}, the membrane potential $u_i$ and output spiking state $o_i$ of IF neuron $i$ in the spiking convolutional layer are decided by:
\begin{equation}
	\label{eq:single_sp_conv}
	\eqalign{u_i = Conv(X), \cr
		o_i=\cases{1&for $u_i \geq u_{th}$\\
			0 &for otherwise\\},} 
\end{equation}
where $Conv(X)$ is the weighted convolutional summation of the inputs from previous layers and $t$ in Eq.~\ref{eq:IF} is ignored since the model only updates once. After spiking feature extraction, there are $N_e$ encoder blocks to encode the spiking representation. Each encoder employs a max pooling layer and multiple spiking convolutional layers to capture surface normal features. The neuronal dynamics of IF neurons in these layers are still controlled by Eq.~\ref{eq:single_sp_conv}. The encoded features are subsequently decoded using $N_d$ decoder blocks, where $N_d=N_e$. Since transposed convolutions are often associated with the creation of checkerboard artifacts~\cite{odena2016deconvolution}, each decoder consists of an upsampling layer followed by multiple spiking convolutional layers, where the IF neurons are governed by Eq.~\ref{eq:single_sp_conv}. For the upsampling operations, we have two options: nearest neighbor upsampling and bilinear upsampling. Through our experiments, we will show that nearest neighbor upsampling can achieve performance comparable to bilinear upsampling in event-based surface normal estimation while preserving the fully spiking nature of our proposed model. As suggested in the UNet architecture, to address the challenge of information loss during down-sampling and up-sampling, skip connections are utilized between corresponding encoder and decoder blocks at the same hierarchical levels. To preserve the spiking nature and avoid introducing non-binary values, the proposed model utilizes concatenations as skip connections. Lastly, the final prediction layer employs the potential-assisted IF neurons~\cite{strohmer2021integrating, wu2021liaf} to estimate the surface normals. Unlike traditional IF neurons generate spikes based on Eq.~\ref{eq:single_sp_conv}, the potential-assisted IF neurons are non-spiking neurons which output membrane potential driven by:
\begin{equation}
	\label{eq:pa_single_sp_conv}
	\eqalign{u_i = Conv(X), \cr
		o_i = u_i,} 
\end{equation}
where $o_i$ denotes the real-valued output of the neuron $i$. These potential-assisted dynamics can be extended to both LIF and PLIF neurons, facilitating the construction of a Single-Timestep Spiking UNet using these types of neurons. By producing real-valued membrane potential outputs, potential-assisted neurons retain rich information that enhances surface normal estimation and boosts the expressivity of SNNs, especially for large-scale regression tasks.


\subsubsection{Multi-Timestep Spiking UNet}

To take the advantage of temporal neuronal dynamics of spiking neurons and extract rich temporal information from event-based data, we propose the Multi-Timestep Spiking UNet for event-based shape from polarization. Figure~\ref{multi} shows the network structure of the Multi-Timestep Spiking UNet. Similar to the Single-Timestep Spiking UNet, the Multi-Timestep Spiking UNet also consists of an event encoding module, an encoder, a decoder, and a final surface normal prediction layer. However, unlike the Single-Timestep Spiking UNet processing the CVGR-I representation as a whole and updating the membrane potential of its spiking neurons only once per data sample, the Multi-Timestep Spiking UNet processes the $B\times H\times W$ CVGR-I representation for each data sample along its temporal dimension $B$. At each time step, a $1\times H\times W$ CVGR-I representation is fed in to the event encoding module and transformed as the size of $N_c\times H\times W$, followed by $N_e$ encoder blocks, $N_d$ decoder blocks, and a final prediciton layer. Based on Eq.~\ref{eq:IF}, the membrane potential $u_i(t)$ and output spiking state $o_i(t)$ of IF neuron $i$ in the spiking convolutional layers of the Multi-Timestep Spiking UNet are decided by:
\begin{equation}
	\label{eq:multi_sp_conv}
	\eqalign{u_i(t) = u_i(t - 1)(1 - o_i(t-1)) + Conv(X(t)), \cr
		o_i(t)=\cases{1&for $u_i(t) \geq u_{th}$\\
			0 &for otherwise\\},} 
\end{equation}
where $Conv(X(t))$ is the weighted convolutional summation of the inputs from previous layers at the time step $t$. The final prediction layer continues to use potential-assisted IF neurons, but with temporal dynamics as outlined below:
\begin{equation}
	\label{eq:pa_multi_sp_conv}
	\eqalign{u_i(t) = u_i(t - 1) + Conv(X(t)), \cr
		o_i(t) = u_i(t),} 
\end{equation}
where the potential-assisted IF neuron $i$ accumulates its membrane potential to maintain the rich temporal information, $o_i(t)$ is the output of the neuron $i$ at time step $t$, and we use \textbf{the outputs at the last time step} as the final surface normal predictions.

\subsection{Training and Implementation Details}
We normalize outputs from spiking neurons into unit-length surface normal vectors $\mathbf{\hat{N}}$ and then apply the cosine similarity loss function:
\begin{equation}
	\label{eq:loss}
	\mathcal{L} = \frac{1}{H\times W}\sum_i^H\sum_j^W(1-\left<\mathbf{\hat{N}}_{i,j}, \mathbf{N}_{i,j}\right>),
\end{equation}
where $\left<\cdot,\cdot\right>$ indicates the dot product, $\mathbf{\hat{N}}_{i,j}$ refers to the estimated surface normal at the pixel location $(i, j)$, while $\mathbf{N}_{i,j}$ denotes the ground truth surface normal at the same location. The objective is to minimize this loss, which is achieved when the orientations of $\mathbf{\hat{N}}_{i,j}$ and $\mathbf{N}_{i,j}$ align perfectly.

To optimize the Single-Timestep Spiking UNet, we utilize the backpropagation method~\cite{hecht1992theory} to calculate the weight updates:
\begin{equation}
	\label{eq:w_single}
	\Delta w^l = \frac{\partial\mathcal{L}}{\partial w^l} = \frac{\partial\mathcal{L}}{\partial o^l}\frac{\partial o^l}{\partial u^l}\frac{\partial u^l}{\partial w^l},
\end{equation}
where $w^l$ is the weight for layer $l$, $o^l$ is the output of spiking neurons in layer $l$, and $u^l$ is the membrane potential of spiking neurons in layer $l$. Similarly, to optimize the Multi-Timestep Spiking UNet, we utilize the BackPropagation Through Time (BPTT)~\cite{werbos1990backpropagation} to calculate the weight updates. In BPTT, the model is unrolled for all discrete time steps, and the weight update is computed as the sum of gradients from each time step as follows:
\begin{equation}
	\label{eq:w_multi}
	\Delta w^l = \sum_{t=0}^{B-1}\frac{\partial\mathcal{L}}{\partial o_t^l}\frac{\partial o_t^l}{\partial u_t^l}\frac{\partial u_t^l}{\partial w^l},
\end{equation}
where $w^l$ is the weight for layer $l$, $o_t^l$ is the output of spiking neurons in layer $l$ at the time step $t$, and $u_t^l$ is the membrane potential of spiking neurons in layer $l$ at the time step $t$. Based on the Heaviside step functions in Eq.~\ref{eq:single_sp_conv} and Eq.~\ref{eq:multi_sp_conv}, we can see that both $\frac{\partial o^l}{\partial u^l}$ and $\frac{\partial o_t^l}{\partial u_t^l}$ cannot be differentiable in spiking convolutional layers. To overcome the non-differentiability, we use the differentiable ArcTan function $g(x) = \frac{1}{\pi}arctan(\pi x) + \frac{1}{2}$ as the surrogate funciton of the Heaviside step function~\cite{neftci2019surrogate}. For the final prediction layer with potential-assisted spiking neurons, since they output membrane potential instead of spikes, we have $\frac{\partial o^l}{\partial u^l} = 1$ and $\frac{\partial o_t^l}{\partial u_t^l} = 1$ for these layers' weight updates.

\section{Experiments and Results}
In this section, we evaluate the effectiveness and efficiency of our proposed SNN models on event-based shape from polarization. We begin by introducing the experimental setup, datasets, baselines, and performance metrics for event-based shape from polarization. Then, extensive experiments on these datasets showcase the capabilities of our models, both in quantitative and qualitative terms, across synthetic and real-world scenarios. Lastly, we analyze the computational costs of our models to highlight their enhanced energy efficiency compared to the counterpart ANN models.
\subsection{Experimental Setup}
Our models are implemented with SpikingJelly~\cite{fang2023spikingjelly}, an open-source deep learning framework for SNNs based on PyTorch~\cite{paszke2019pytorch}. To fairly compare with the counterpart ANN models, we ensure our models have the similar settings like the ANN models in~\cite{muglikar2023event}. Specifically, we set $B=8$ for the input event representation. In addition, our models have $N_e=4$ encoder blocks and $N_d=4$ decoder blocks. And the event encoding module outputs the binary spiking representation with the channel size of $N_c=64$. For the spiking-related settings, all the spiking neurons in the spiking convolutional layers are set with a reset value ($u_{reset}$) of 0 and a threshold value ($u_{th}$) of $1$. Following~\cite{fang2021incorporating, ledinauskas2020training}, normalization techniques are applied after each convolution (Conv) operation for faster convergence. We train our models for $1000$ epochs with a batch size of 2 on Quadro RTX 8000. We use the Adam~\cite{kingma2014adam} with a learning rate of $1e-4$ to optimize our models.

\subsection{Datasets}
We evaluate our proposed models on two latest large-scale datasets for event-based shape from polarization, including the ESfP-Synthetic Dataset and ESfP-Real Dataset. 

The ESfP-Synthetic Dataset was generated using the Mitsuba renderer~\cite{Mitsuba3}, which created scenes with textured meshes illuminated by a point light source. For each scene, a polarizer lens, positioned in front of the camera, was rotated through angles ranging from 0 to 180 degrees with 15 degrees intervals, producing a total of 12 polarization images. With these images, events were simulated using ESIM~\cite{rebecq2018esim} with a 5\% contrast threshold. Therefore, each scene in the dataset is accompanied by rendered polarization images, simulated events, and groundtruth surface normals provided by the renderer. 

The ESfP-Real Dataset is the first large-scale real-world dataset for event-based shape from polarization. It contains various scenes with different objects, textures, shapes, illuminations, and scene depths. The dataset was collected using a Prophesee Gen 4 event camera~\cite{finateu20201280x720}, a Breakthrough Photography X4 CPL linear polarizer~\cite{X4CPL}, a Lucid Polarisens camera~\cite{LucidV}, and a laser point projector. Specificially, the polarizer rotated in front of the event camera that captured the events for each scene in the dataset. The Lucid Polarisens camera was used to collect polarization images of the same scene at 4 polarization angles \{0, 45, 90, 135\}. And the groundtruth surface normals were generated using Event-based Structured Light~\cite{muglikar2021esl}, a technique that involves integrating the laser point projector with the event camera. 

\subsection{Baselines and Performance Metrics}

We evaluate our models against the state-of-the-art physics-based and learning-based methods in the field of shape from polarization. Smith \textit{et al.}~\cite{smith2018height} combine the physics-based shape from polarization with the photometric image formation model. The method directly estimates lighting information and calculates the surface height using a single polarization image under unknown illumination. Mahmoud \textit{et al.}~\cite{mahmoud2012direct} present a physics-based method to conduct shape recovery using both polarization and shading information. Recently, Muglikar \textit{et al.}~\cite{muglikar2023event} are pioneers in addressing event-based shape from polarization, employing both physics-based and learning-based approaches. Their models are notable for directly using event data as inputs. In this paper, our focus is on comparing our proposed models with the learning-based model developed by Muglikar \textit{et al.} We aim to demonstrate that our SNN-based models can match their performance while offering greater energy efficiency.

To evaluate the accuracy of the predicted surface normals, we employ four metrics: Mean Angular Error (MAE), \% Angular Error under 11.25 degrees (AE$<$11.25), \% Angular Error under 22.5 degrees (AE$<$22.5), and \% Angular Error under 30 degrees (AE$<$30). MAE is a commonly used metric that quantifies the angular error of the predicted surface normal, where a lower value indicates better performance~\cite{ba2020deep, lei2022shape}. The latter three metrics, collectively referred to as angular accuracy, assess the proportion of pixels with angular errors less than 11.25, 22.5, and 30 degrees, respectively, with higher percentages indicating better accuracy~\cite{muglikar2023event}.

\subsection{Performance on ESfP-Synthetic}

\begin{table}[]
	\caption{\label{t1}Shape from polarization performance on the ESfP-Synthetic Dataset in terms of Mean Angular Error (MAE) and the percentage of pixels under specific angular errors (AE$<\cdot$). The ``Input" column specifies whether the method utilizes events (E) or polarization images (I). E+I[0] means the CVGR-I representation. ``Single" is for the Single-Timestep Spiking UNet. ``Multi'' is for the Multi-Timestep Spiking UNet. ``Bilinear'' and ``Nearest'' represent the bilinear upsampling and nearest neighbor upsampling, respectively. We highlight the top performance in bold, and underline the second-best results. }
	\begin{tabular}{l|c|c|c|c|c|c}
		\hline
		Method           & Input  & Task           & MAE$\downarrow$             & AE\textless{}11.25$\uparrow$ & AE\textless{}22.5$\uparrow$ & AE\textless{}30$\uparrow$ \\ \hline
		Mahmoud \textit{et al.}~\cite{mahmoud2012direct}         &  I    & Physics  & 80.923          & 0.034              & 0.065             & 0.085           \\ \hline
		Smith \textit{et al.}~\cite{smith2018height}            & I    & Physics  & 67.684          & 0.010              & 0.047             & 0.106           \\ \hline
		Muglikar \textit{et al.}~\cite{muglikar2023event}            & E   & Physics & 58.196 & 0.007    & 0.046    & 0.095  \\ \hline
		Muglikar \textit{et al.}~\cite{muglikar2023event}         & E+I[0] & Learning & \bfseries{\small 27.953}          & \bfseries{\small 0.263 }             & \bfseries{\small 0.527 }            & \bfseries{\small 0.655}         \\ \hline
		Single\_Bilinear & E+I[0] & Learning & 36.432          & 0.181              & 0.403             & 0.525           \\ \hline
		Single\_Nearest  & E+I[0] & Learning & 36.824          & 0.141              & 0.370             & 0.491           \\ \hline
		Multi\_Bilinear  & E+I[0] & Learning & \underline{31.296}    & \underline{0.200}        & \underline{0.438}       & \underline{0.578}     \\ \hline
		Multi\_Nearest   & E+I[0] & Learning & 31.724          & 0.193              & 0.425             & 0.562           \\ \hline
	\end{tabular}
\end{table}

\begin{table}[]
	\caption{\label{t2}Ablation study on various spiking neurons.}
	\begin{tabular}{l|c|c|c|c|c|c}
		\hline
		Method           & Input  & Task           & MAE$\downarrow$             & AE\textless{}11.25$\uparrow$ & AE\textless{}22.5$\uparrow$ & AE\textless{}30$\uparrow$ \\ \hline
		Multi\_Nearest\_IF   & E+I[0] & Learning & 31.724          & 0.193              & 0.425             & 0.562           \\ \hline
		Multi\_Nearest\_LIF  & E+I[0] & Learning & 35.250
          & 0.154
           & 0.384                 & 0.523           \\ \hline
		Multi\_Nearest\_PLIF  & E+I[0] & Learning & 35.086
   &0.154       & 0.393     &0.530   \\ \hline
	\end{tabular}
\end{table}

\begin{figure}[htp]
	
	\centering
	
	\begin{tabular}{
			@{}
			*{6}{c@{\hspace{4pt}}} 
			c
			@{}
		}
		\column{
			\begin{overpic}[width=2.1cm, percent]{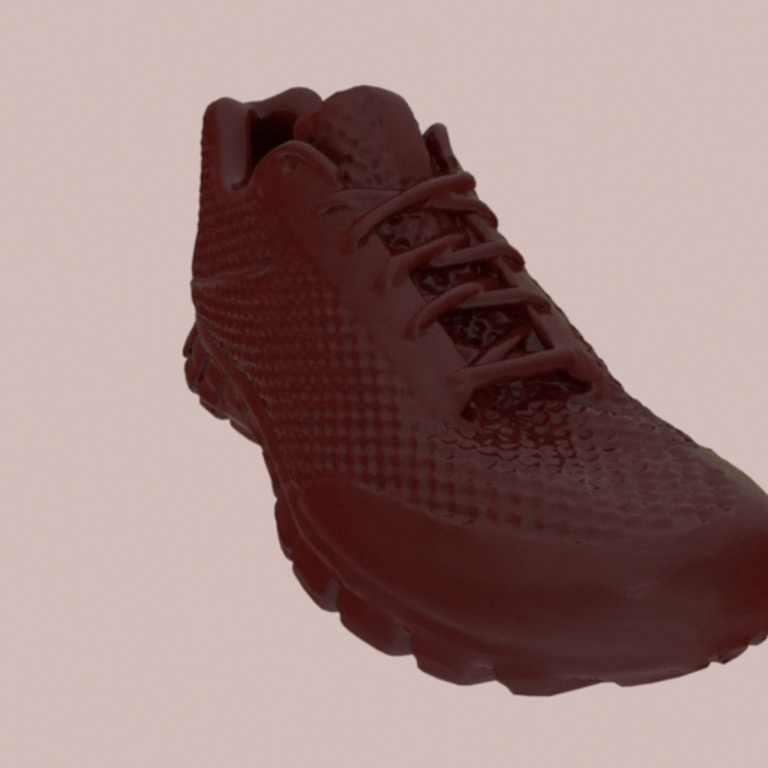}\end{overpic}\\
			\begin{overpic}[width=2.1cm, percent]{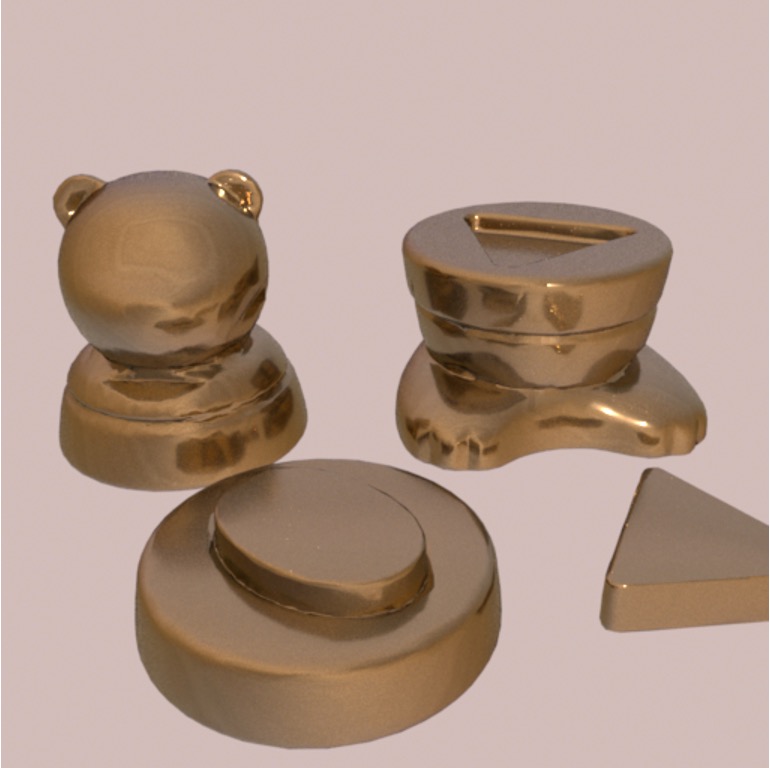}\end{overpic}\\
			\begin{overpic}[width=2.1cm, percent]{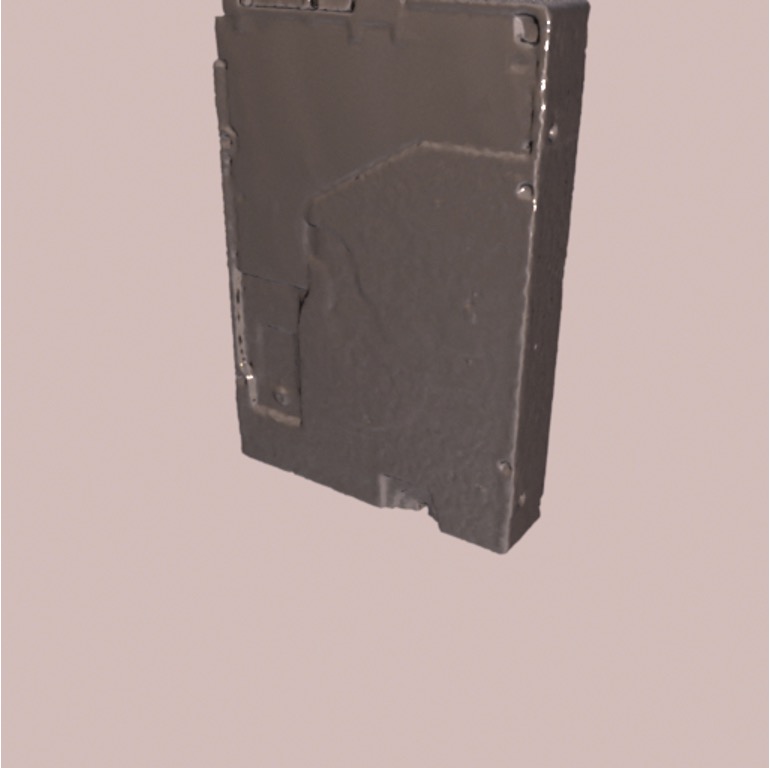}\end{overpic}\\
			\begin{overpic}[width=2.1cm, percent]{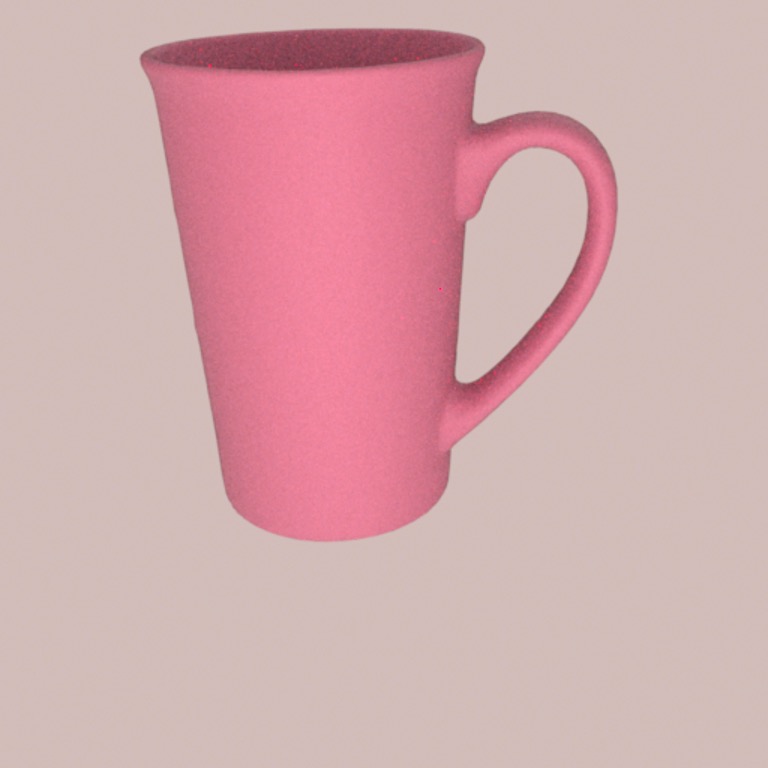}\end{overpic}\\
			\begin{overpic}[width=2.1cm, percent]{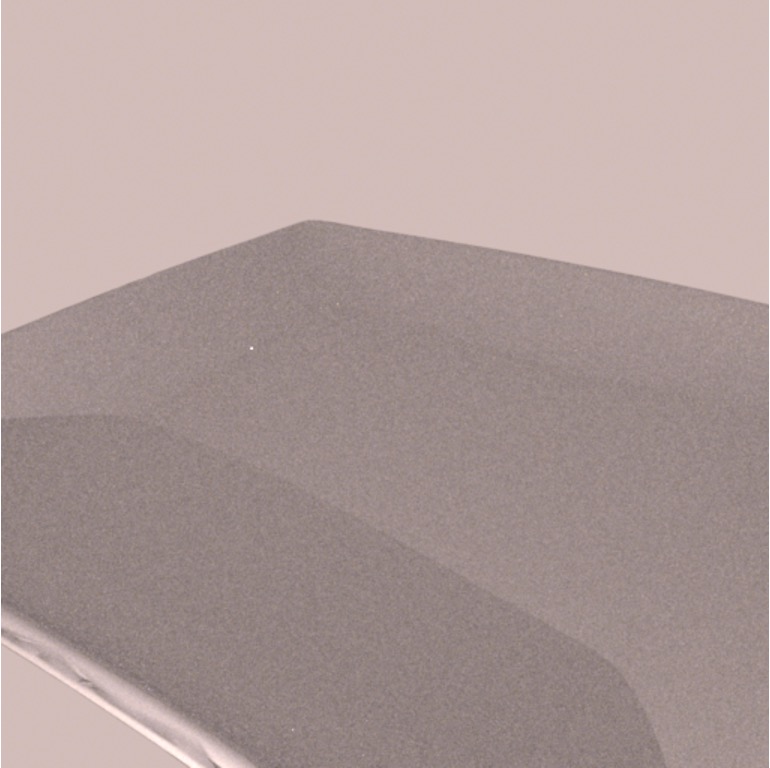}\end{overpic}\\
			\begin{overpic}[width=2.1cm, percent]{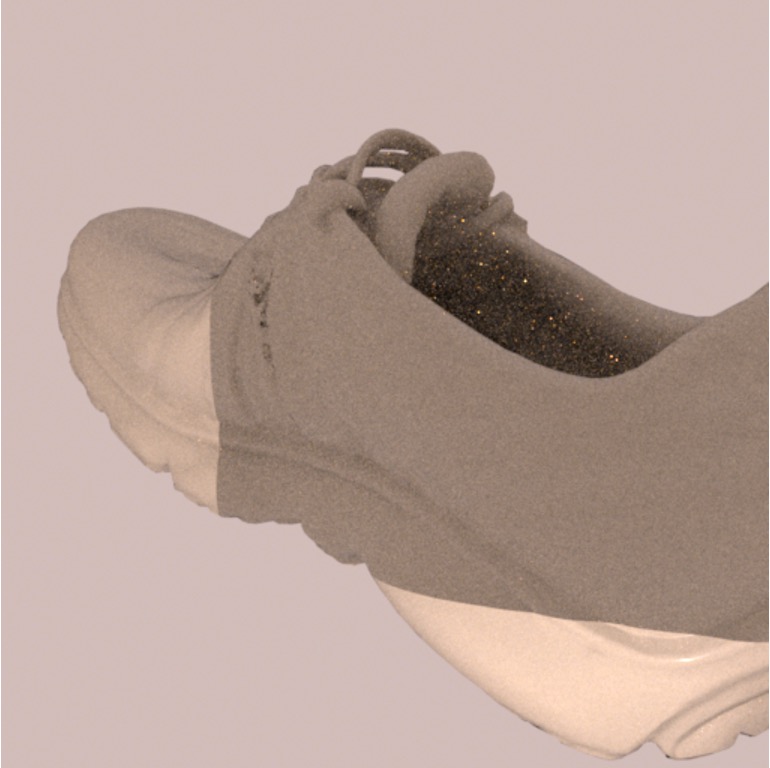}\end{overpic}\\
			\begin{overpic}[width=2.1cm, percent]{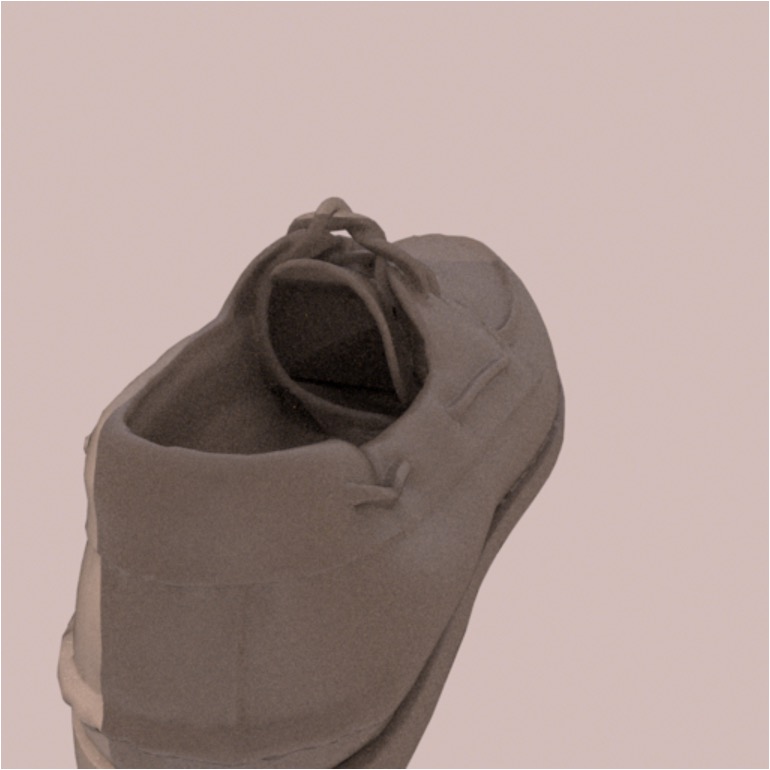}\end{overpic}\\
			\begin{overpic}[width=2.1cm, percent]{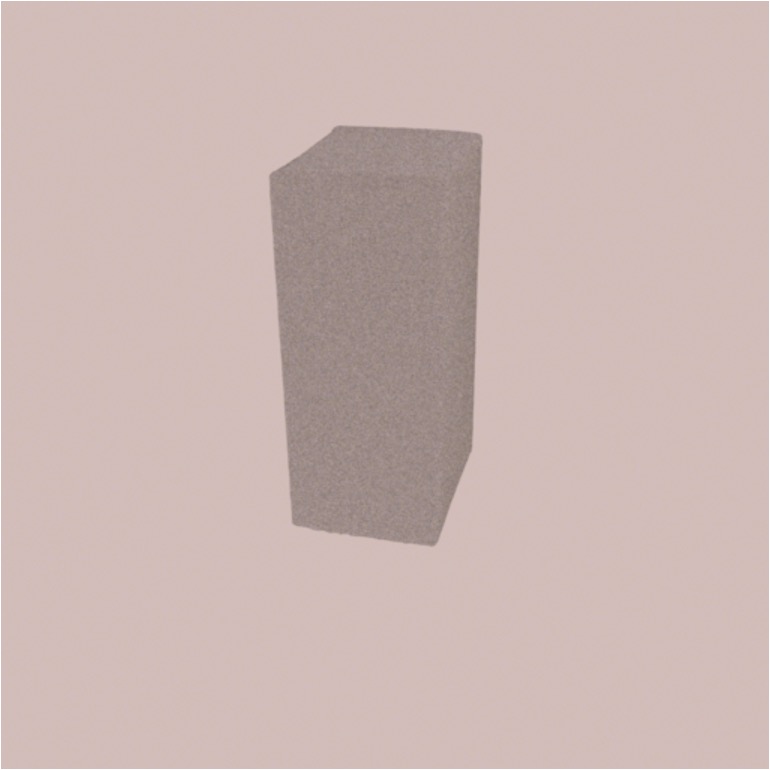}\end{overpic}
		}{Scene\\(a)}
		&
		\column{
			\begin{overpic}[width=2.1cm, percent]{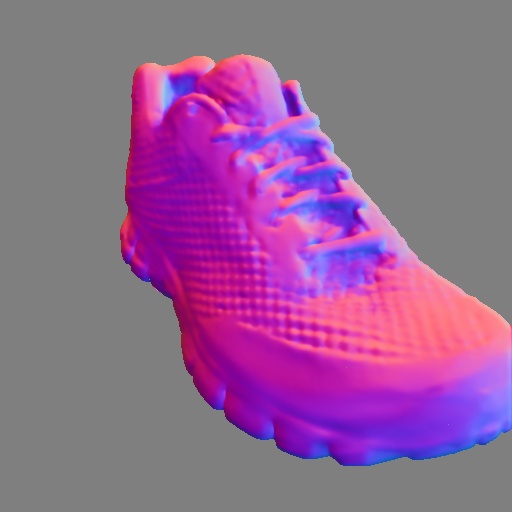}\put(0,85){\colorbox{green}{\color{blue} \small 22.93}}\end{overpic}\\
			\begin{overpic}[width=2.1cm, percent]{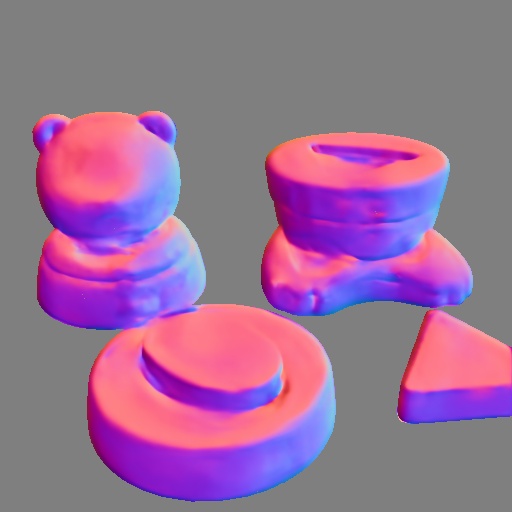}\put(0,85){\colorbox{green}{\color{blue} \small 19.21}}\end{overpic}\\
			\begin{overpic}[width=2.1cm, percent]{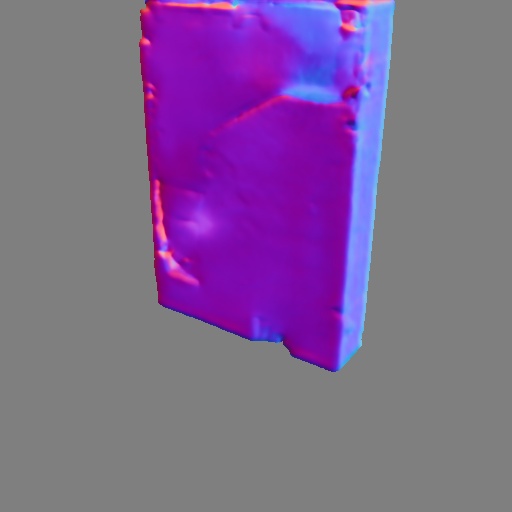}\put(0,85){\color{blue} \colorbox{green}{\small 19.28}}\end{overpic}\\
			\begin{overpic}[width=2.1cm, percent]{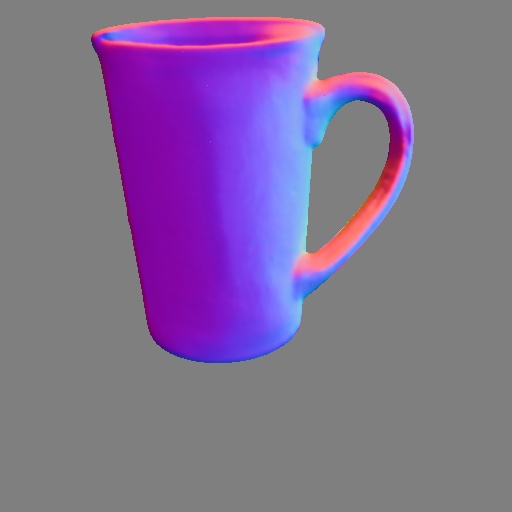}\put(0,85){\colorbox{green}{\color{blue} \small 19.06}}\end{overpic}\\
			\begin{overpic}[width=2.1cm, percent]{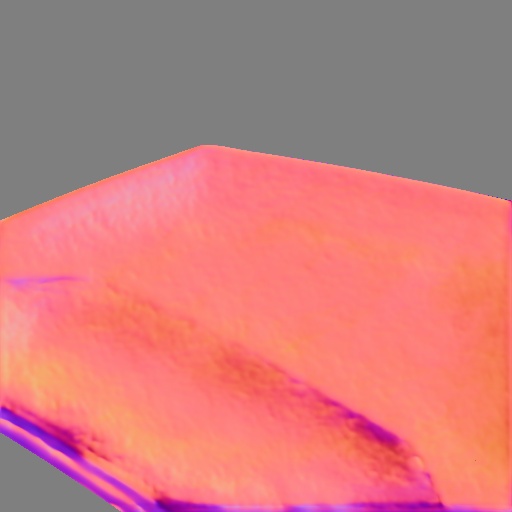}\put(0,85){\color{blue} \small 18.08}\end{overpic}\\
			\begin{overpic}[width=2.1cm, percent]{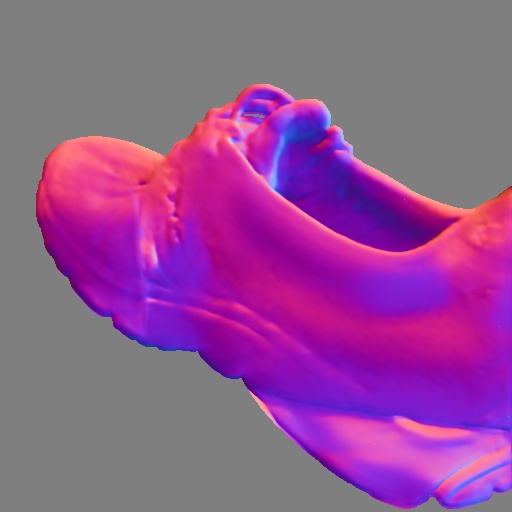}\put(0,85){\color{blue} \small 28.12}\end{overpic}\\
			\begin{overpic}[width=2.1cm, percent]{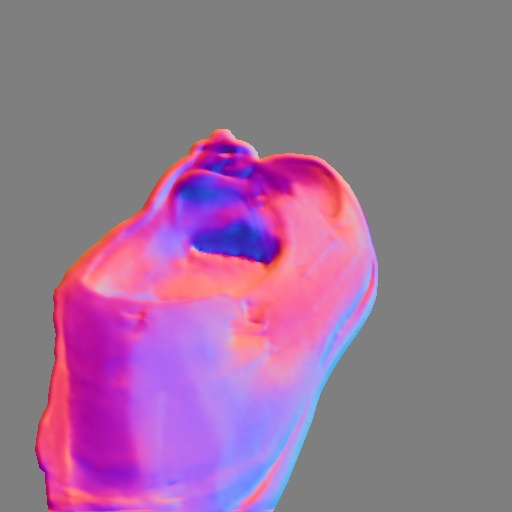}\put(0,85){\color{blue} \small 31.63}\end{overpic}\\
			\begin{overpic}[width=2.1cm, percent]{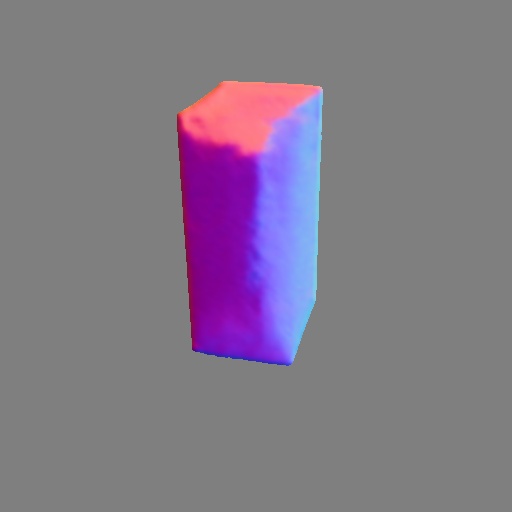}\put(0,85){\color{blue} \small 30.13}\end{overpic}
		}{ANNs~\cite{muglikar2023event}\\(b)}
		&
		\column{
			\begin{overpic}[width=2.1cm, percent]{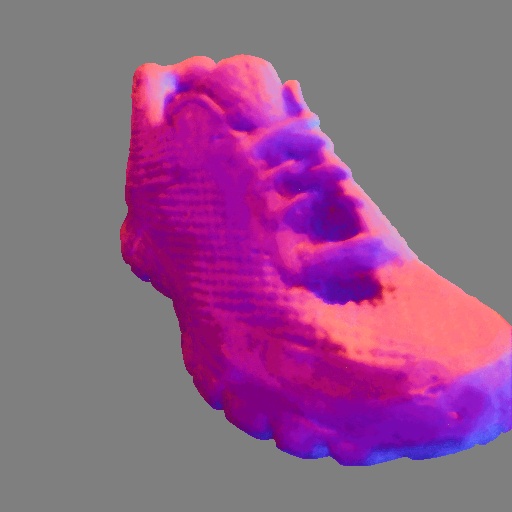}\put(0,85){\color{blue} \small 25.68}\end{overpic}\\
			\begin{overpic}[width=2.1cm, percent]{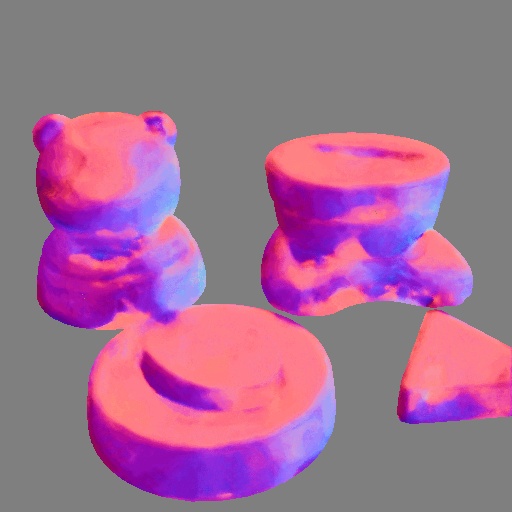}\put(0,85){\color{blue} \small 25.23}\end{overpic}\\
			\begin{overpic}[width=2.1cm, percent]{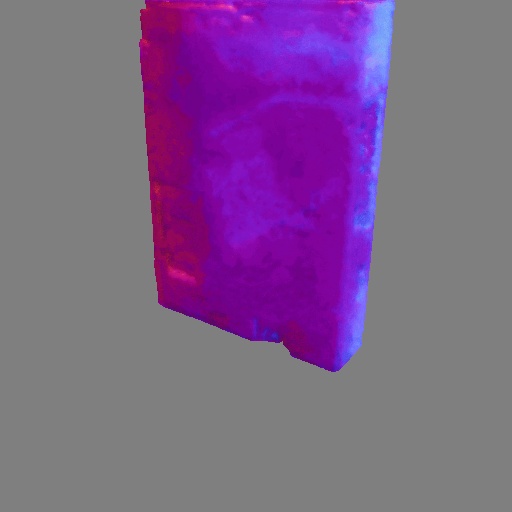}\put(0,85){\color{blue} \small 20.25}\end{overpic}\\
			\begin{overpic}[width=2.1cm, percent]{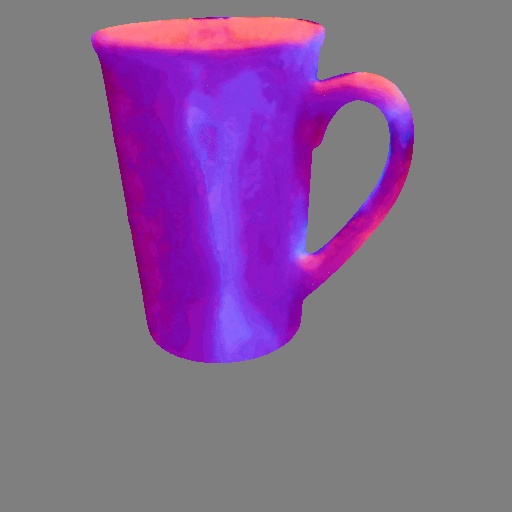}\put(0,85){\color{blue} \small 31.81}\end{overpic}\\
			\begin{overpic}[width=2.1cm, percent]{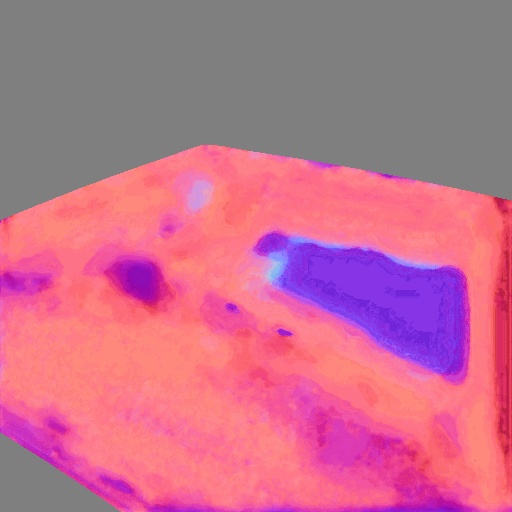}\put(0,85){\color{blue} \small 30.58}\end{overpic}\\
			\begin{overpic}[width=2.1cm, percent]{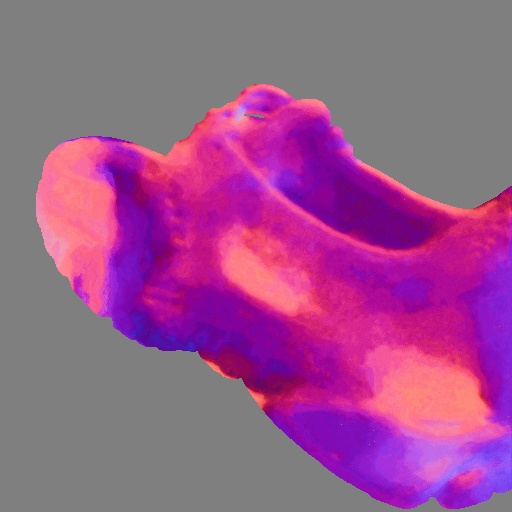}\put(0,85){\color{blue} \small 37.82}\end{overpic}\\
			\begin{overpic}[width=2.1cm, percent]{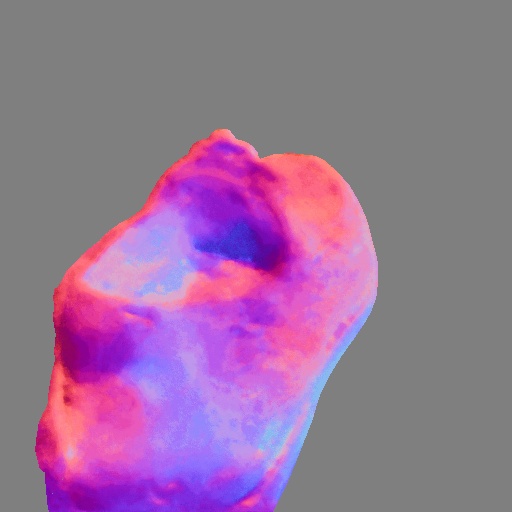}\put(0,85){\color{blue} \small 34.71}\end{overpic}\\
			\begin{overpic}[width=2.1cm, percent]{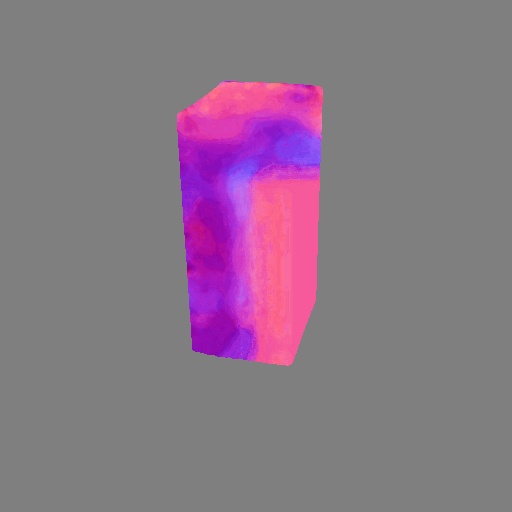}\put(0,85){\color{blue} \small 46.01}\end{overpic}
		}{Single\_B\\(c)}
		&
		\column{
			\begin{overpic}[width=2.1cm, percent]{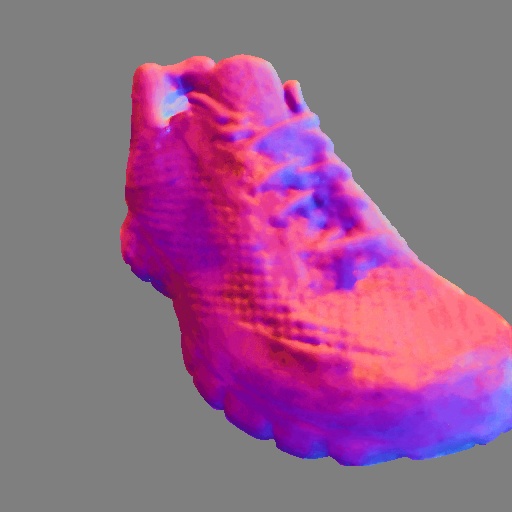}\put(0,85){\color{blue} \small 23.08}\end{overpic}\\
			\begin{overpic}[width=2.1cm, percent]{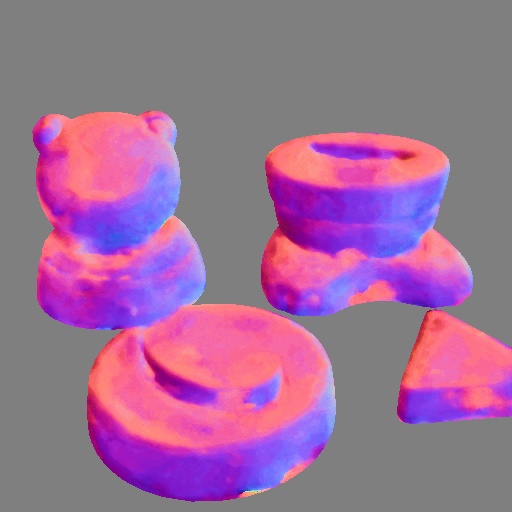}\put(0,85){\color{blue} \small 23.63}\end{overpic}\\
			\begin{overpic}[width=2.1cm, percent]{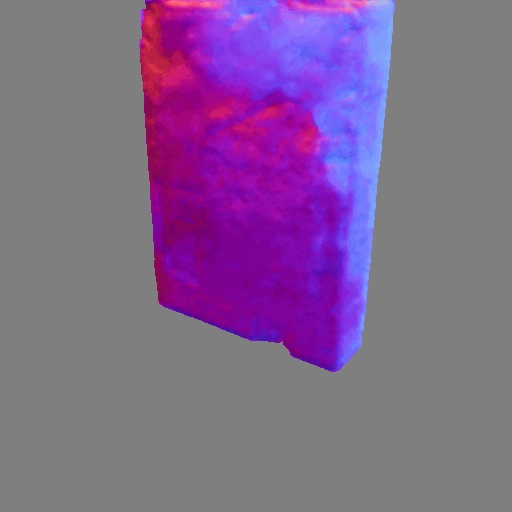}\put(0,85){\color{blue} \small 28.12}\end{overpic}\\
			\begin{overpic}[width=2.1cm, percent]{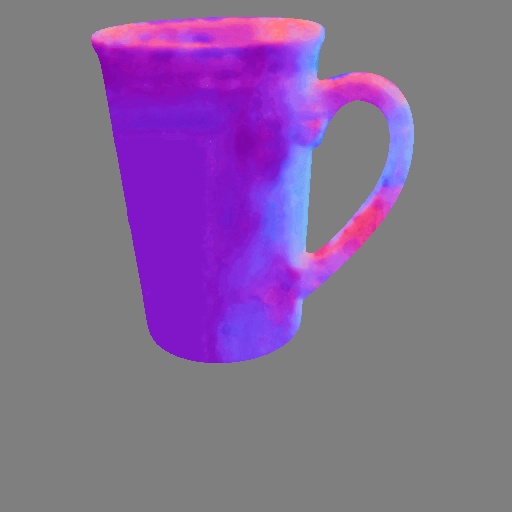}\put(0,85){\color{blue} \small 30.67}\end{overpic}\\
			\begin{overpic}[width=2.1cm, percent]{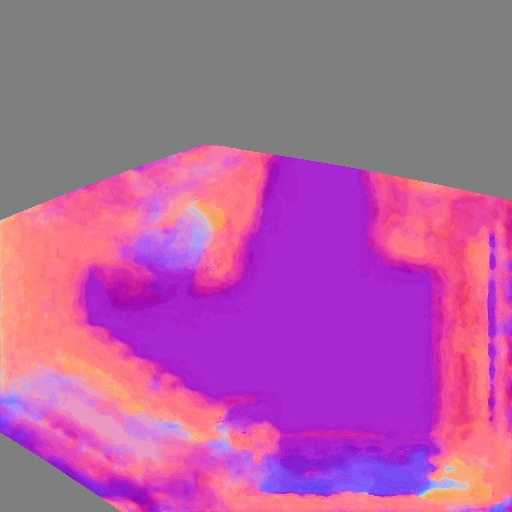}\put(0,85){\color{blue} \small 48.09}\end{overpic}\\
			\begin{overpic}[width=2.1cm, percent]{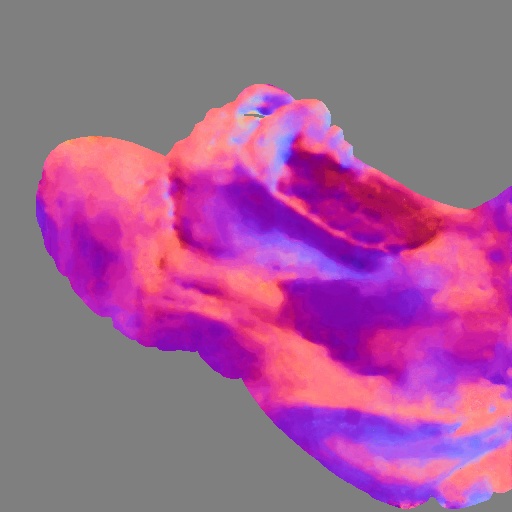}\put(0,85){\color{blue} \small 44.53}\end{overpic}\\
			\begin{overpic}[width=2.1cm, percent]{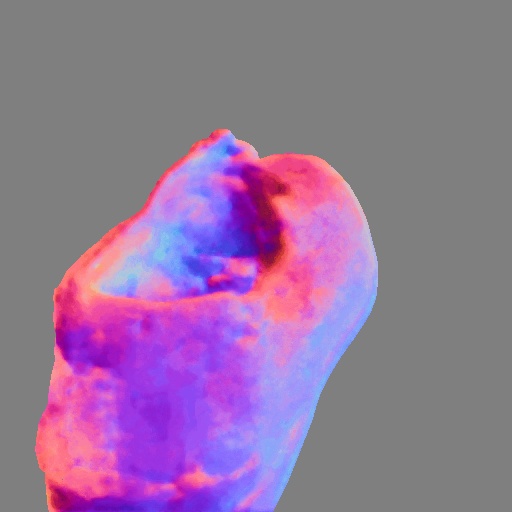}\put(0,85){\color{blue} \small 34.17}\end{overpic}\\
			\begin{overpic}[width=2.1cm, percent]{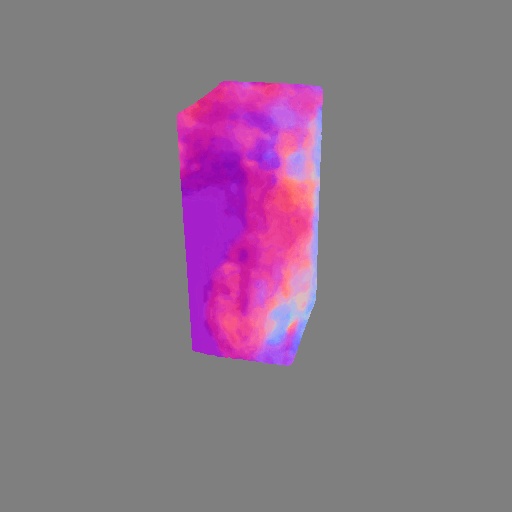}\put(0,85){\color{blue} \small 45.21}\end{overpic}
		}{Single\_N\\(d)}
		&
		\column{
			\begin{overpic}[width=2.1cm, percent]{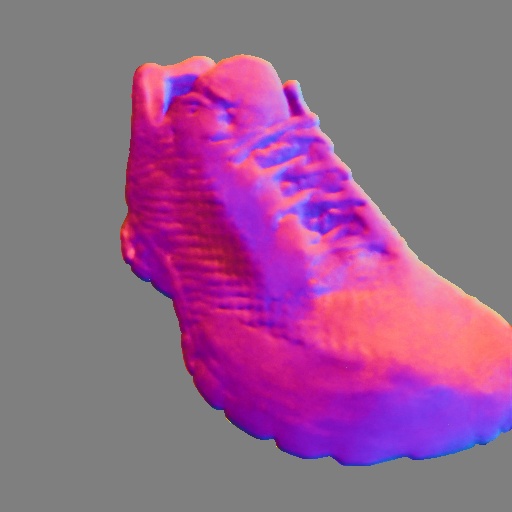}\put(0,85){\color{blue} \small 25.11}\end{overpic}\\
			\begin{overpic}[width=2.1cm, percent]{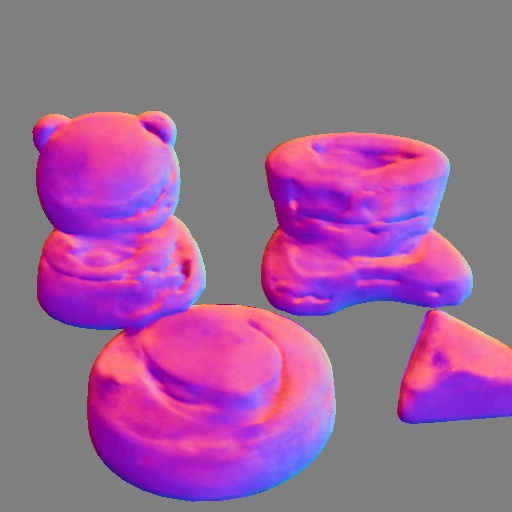}\put(0,85){\color{blue} \small 29.00}\end{overpic}\\
			\begin{overpic}[width=2.1cm, percent]{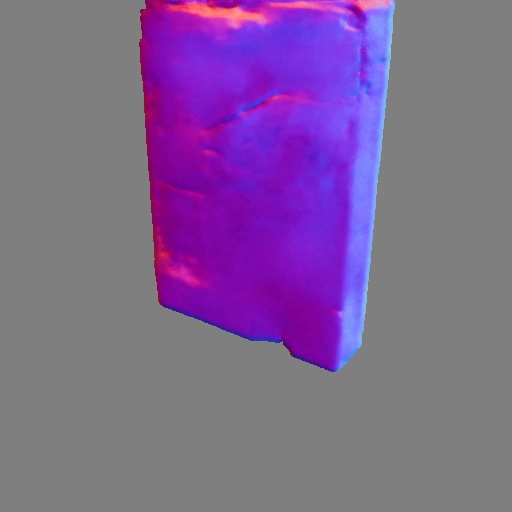}\put(0,85){\color{blue} \small 24.90}\end{overpic}\\
			\begin{overpic}[width=2.1cm, percent]{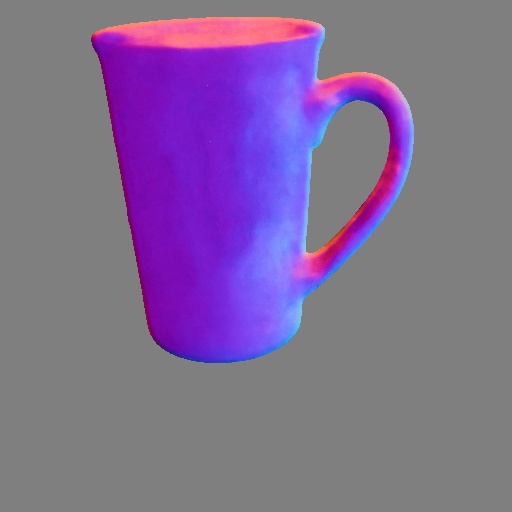}\put(0,85){\color{blue} \small 20.98}\end{overpic}\\
			\begin{overpic}[width=2.1cm, percent]{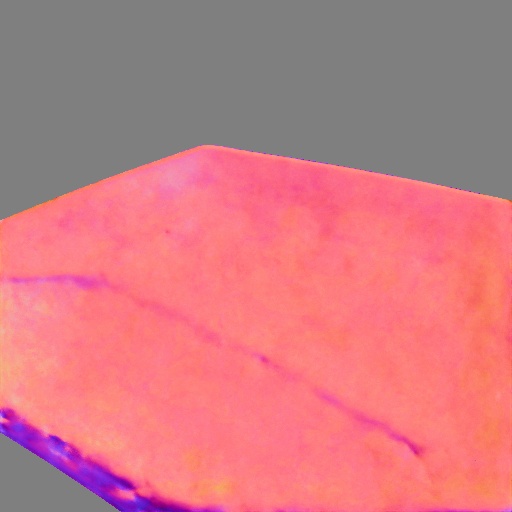}\put(0,85){\colorbox{green}{\color{blue} \small 15.89}}\end{overpic}\\
			\begin{overpic}[width=2.1cm, percent]{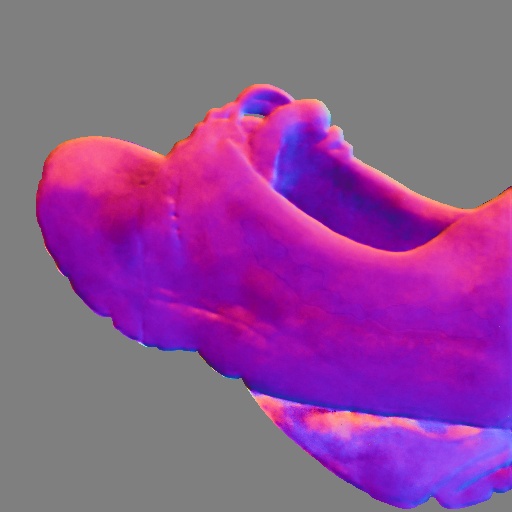}\put(0,85){\color{blue} \small 27.30}\end{overpic}\\
			\begin{overpic}[width=2.1cm, percent]{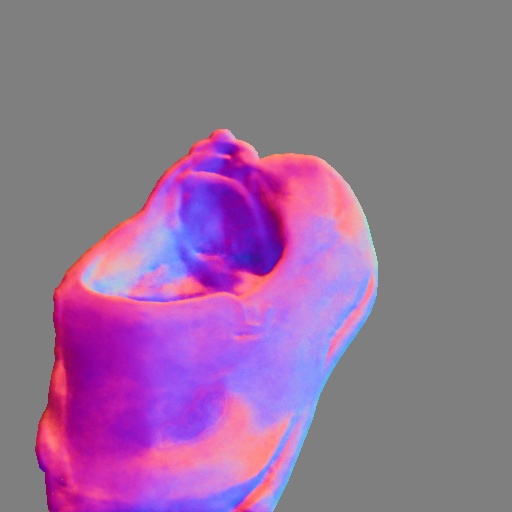}\put(0,85){\color{blue} \small 32.36}\end{overpic}\\
			\begin{overpic}[width=2.1cm, percent]{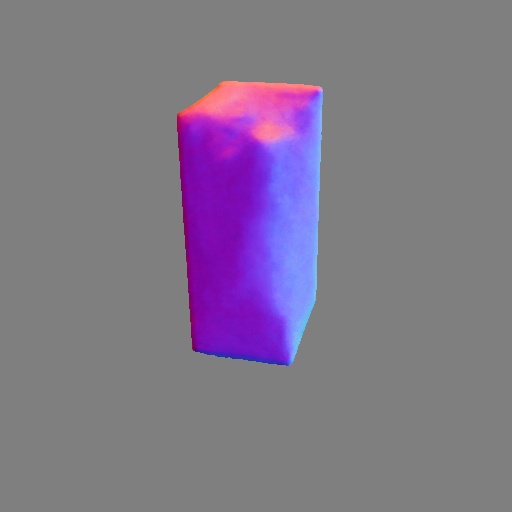}\put(0,85){\colorbox{green}{\color{blue} \small 23.77}}\end{overpic}
		}{Multi\_B\\(e)}
		&
		\column{
			\begin{overpic}[width=2.1cm, percent]{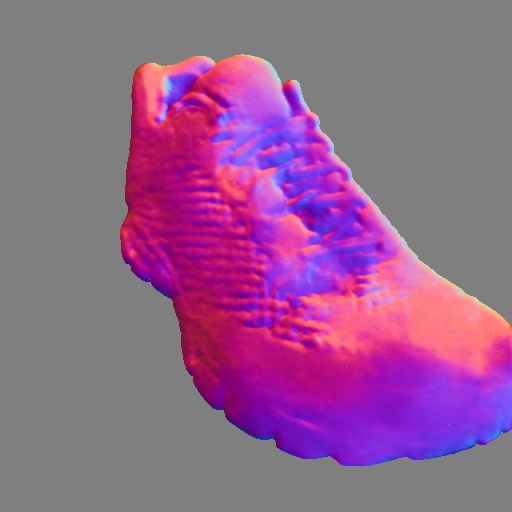}\put(0,85){\color{blue} \small 26.52}\end{overpic}\\
			\begin{overpic}[width=2.1cm, percent]{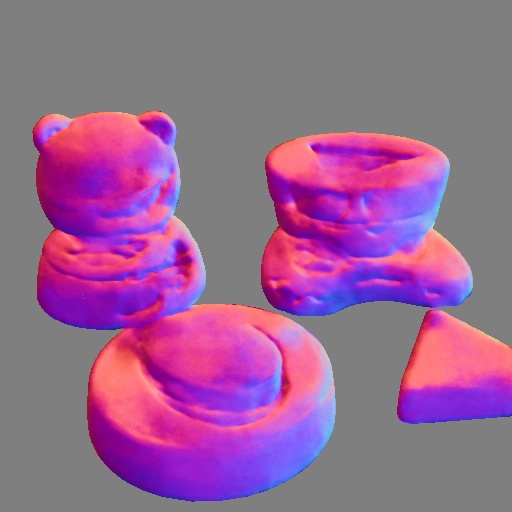}\put(0,85){\color{blue} \small 29.19}\end{overpic}\\
			\begin{overpic}[width=2.1cm, percent]{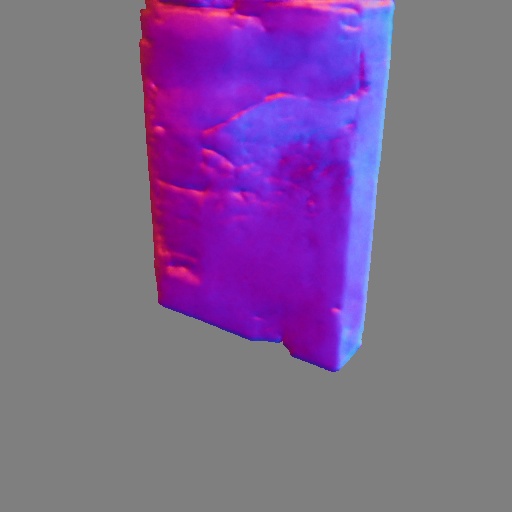}\put(0,85){\color{blue} \small 25.44}\end{overpic}\\
			\begin{overpic}[width=2.1cm, percent]{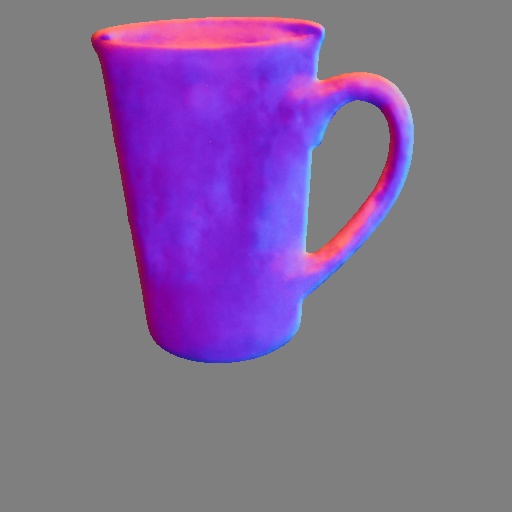}\put(0,85){\color{blue} \small 25.38}\end{overpic}\\
			\begin{overpic}[width=2.1cm, percent]{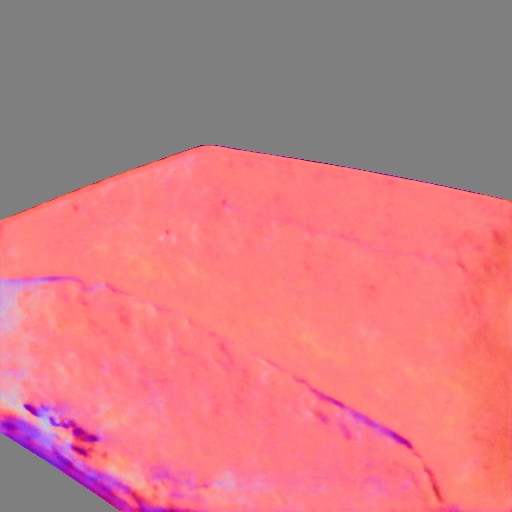}\put(0,85){\color{blue} \small 17.80}\end{overpic}\\
			\begin{overpic}[width=2.1cm, percent]{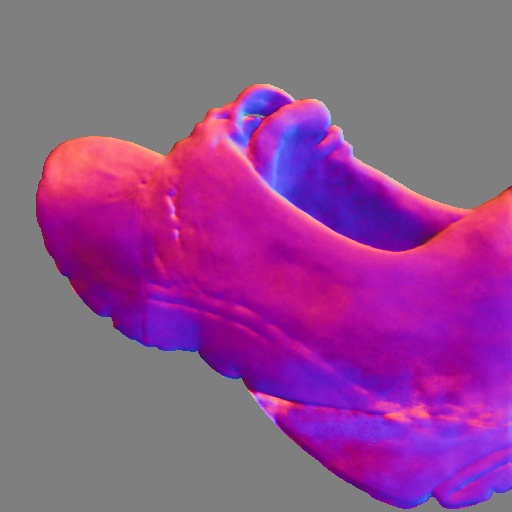}\put(0,85){\colorbox{green}{\color{blue} \small 25.27}}\end{overpic}\\
			\begin{overpic}[width=2.1cm, percent]{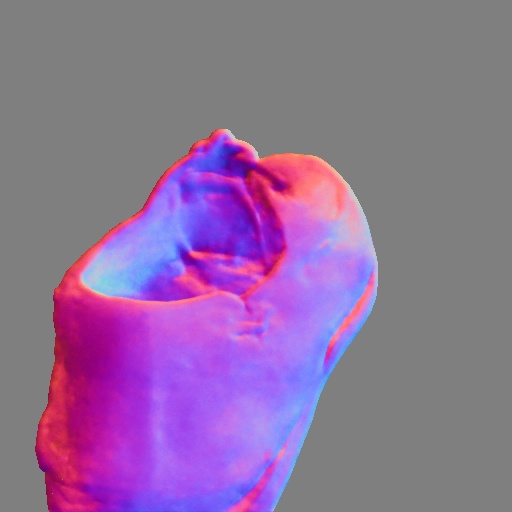}\put(0,85){\colorbox{green}{\color{blue} \small 30.18}}\end{overpic}\\
			\begin{overpic}[width=2.1cm, percent]{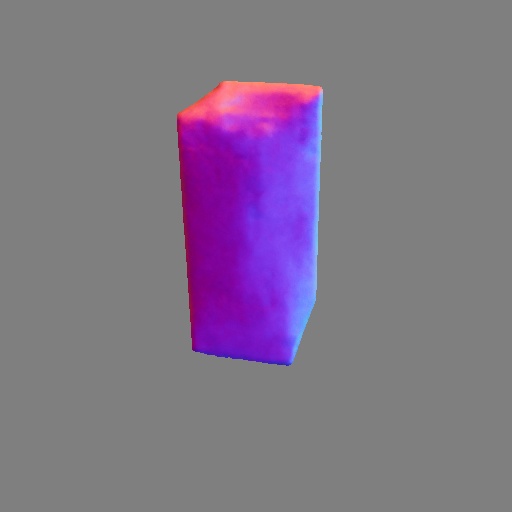}\put(0,85){\color{blue} \small 25.30}\end{overpic}
		}{Multi\_N\\(f)}
		&
		\column{
			\begin{overpic}[width=2.1cm, percent]{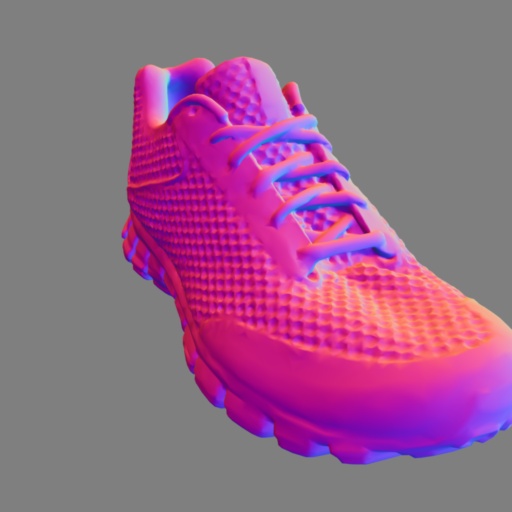}\end{overpic}\\
			\begin{overpic}[width=2.1cm, percent]{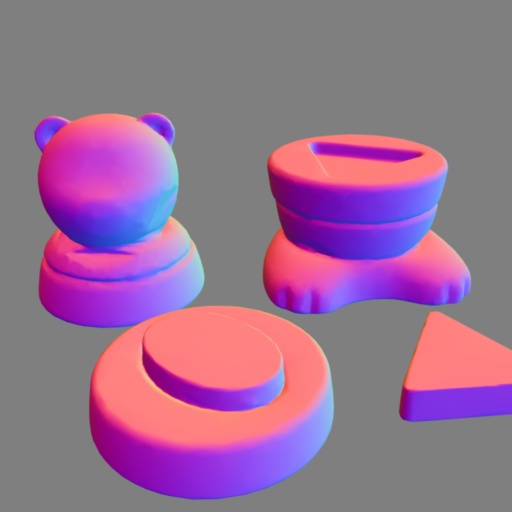}\end{overpic}\\
			\begin{overpic}[width=2.1cm, percent]{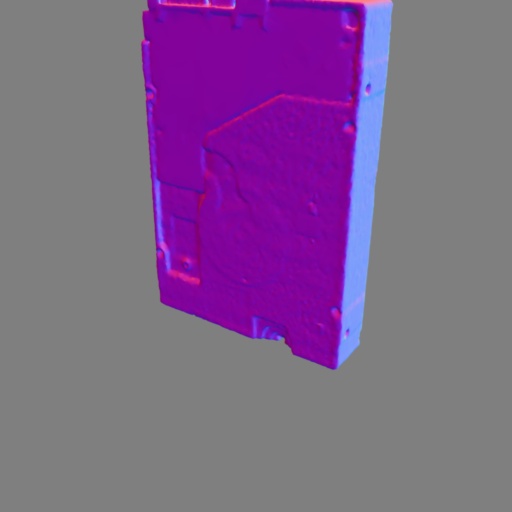}\end{overpic}\\
			\begin{overpic}[width=2.1cm, percent]{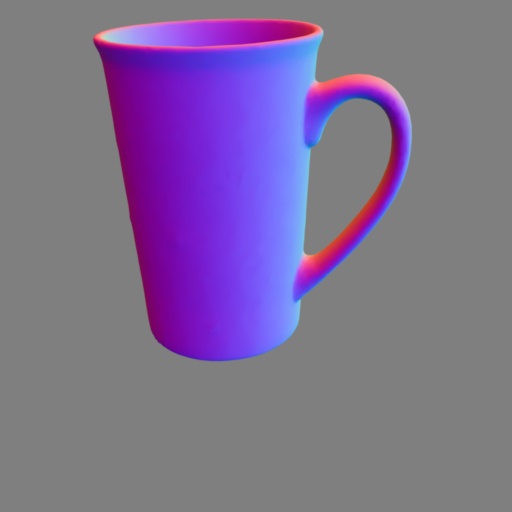}\end{overpic}\\
			\begin{overpic}[width=2.1cm, percent]{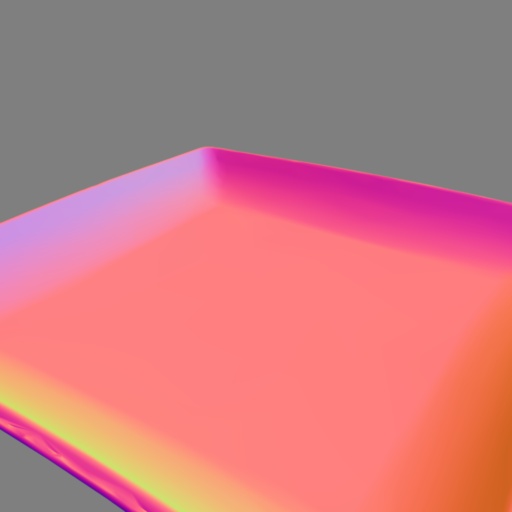}\end{overpic}\\
			\begin{overpic}[width=2.1cm, percent]{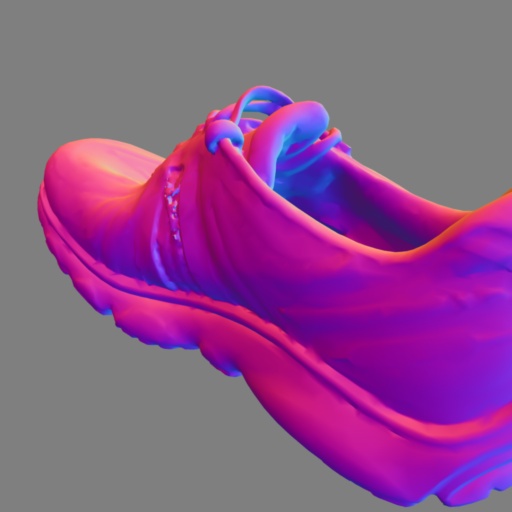}\end{overpic}\\
			\begin{overpic}[width=2.1cm, percent]{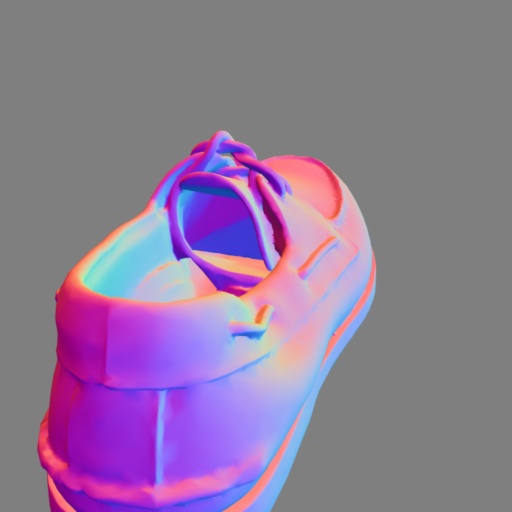}\end{overpic}\\
			\begin{overpic}[width=2.1cm, percent]{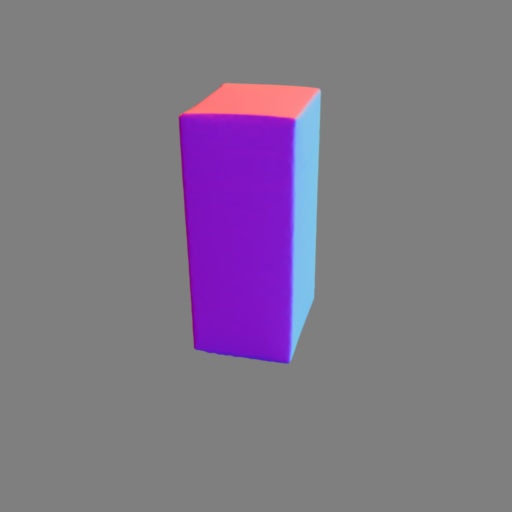}\end{overpic}
		}{GT\\(g)}
	\end{tabular}
	
	\caption{\label{exp1}Qualitative results on the ESfP-Synthetic Dataset. Column (a) shows the scene photographs for context. Column (b) is for the counterpart ANN models. Columns (c-d) are for the Single-Timestep Spiking UNets with bilinear upsampling and nearest neighbor upsampling, respectively. Columns (e-f) are for the Multi-Timestep Spiking UNets with bilinear upsampling and nearest neighbor upsampling, respectively. Column (g) presents the ground truth normals. The MAE for the reconstructions is shown on the top left of each cell in Columns (b-f). For each scene, we highlight the best result using the green colorbox.}
	
\end{figure}

We thoroughly evaluate our proposed models on the ESfP-Synthetic Dataset, using both quantitative metrics and qualitative analysis. Specifically, Table~\ref{t1} presents the performance of both baselines and our methods in surface normal estimation on the ESfP-Synthetic Dataset. In addition, Figure~\ref{exp1} showcases the qualitative results of our models and the ANN counterpart on the ESfP-Synthetic Dataset.

From Table~\ref{t1}, we can see that our proposed models significantly outperform the physics-based methods. The reason why our model can achieve the better performance is that our models benefit from the large-scale dataset and utilize the spiking neurons to extract useful information for event-based shape from polarization. Despite this success, our models do not quite match the \textbf{overall performance} of their ANN counterpart on this dataset, likely due to the limited representation capacity of spiking neurons. However, as Fig.~\ref{exp1} illustrates, our Multi-Timestep Spiking UNets still manage to \textbf{achieve comparable, and in some cases superior, results in shape recovery across various objects in the test set}, compared to the ANN models. 

Table~\ref{t1} clearly demonstrates that the temporal dynamics inherent in spiking neurons enable the Multi-Timestep Spiking UNets to surpass the Single-Timestep versions in surface normal estimation. Additionally, nearest neighbor sampling, as compared to bilinear upsampling, shows comparable performance while preserving the binary nature and compatibility with SNNs.

Recognizing the effectiveness of Multi-Timestep Spiking UNets, we undertook an ablation study aimed at identifying the ideal spiking neurons to fully leverage their temporal dynamic capabilities. The results, detailed in Table~\ref{t2}, indicate that IF neurons offer superior performance. This is largely due to their ability to retain more extensive temporal information, as they operate without the influence of a leaky factor.

\subsection{Performance on ESfP-Real}
\begin{figure}[htp]
	\centering
	
	\begin{tabular}{
			@{}
			*{6}{c@{\hspace{4pt}}} 
			c
			@{}
		}
		\column{
			\begin{overpic}[width=2.1cm, percent]{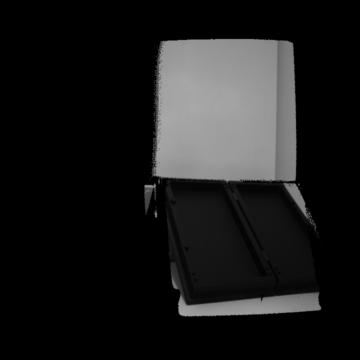}\end{overpic}\\
			\begin{overpic}[width=2.1cm, percent]{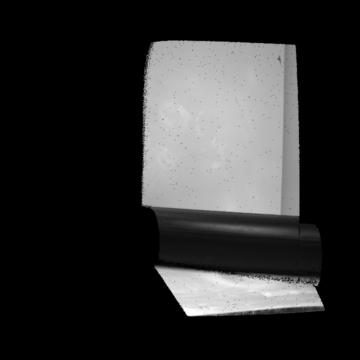}\end{overpic}\\
			\begin{overpic}[width=2.1cm, percent]{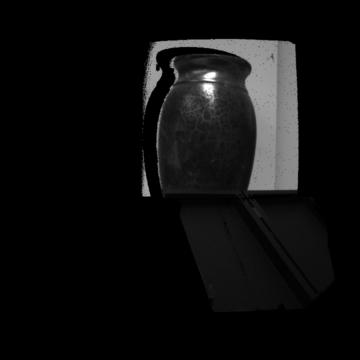}\end{overpic}\\
			\begin{overpic}[width=2.1cm, percent]{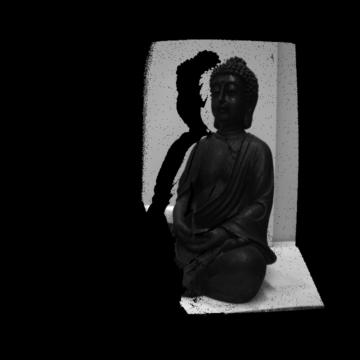}\end{overpic}\\
			\begin{overpic}[width=2.1cm, percent]{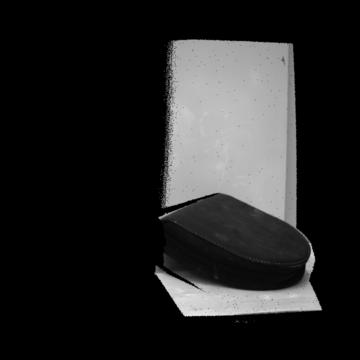}\end{overpic}\\
			\begin{overpic}[width=2.1cm, percent]{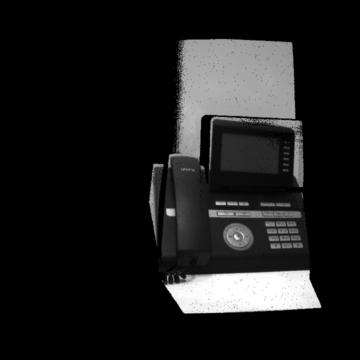}\end{overpic}\\
			\begin{overpic}[width=2.1cm, percent]{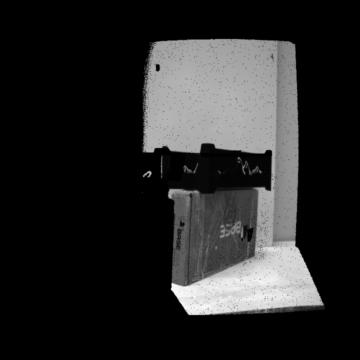}\end{overpic}\\
			\begin{overpic}[width=2.1cm, percent]{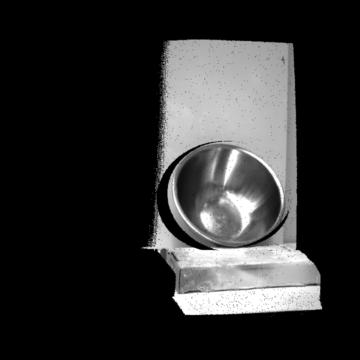}\end{overpic}
		}{Scene\\(a)}
		&
		\column{
			\begin{overpic}[width=2.1cm, percent]{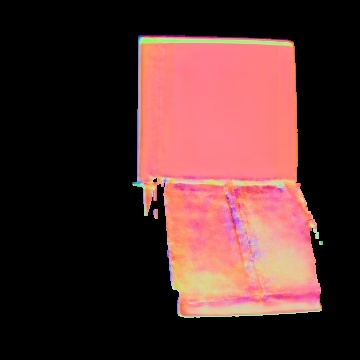}\put(0,85){\color{white} \small 28.36}\end{overpic}\\
			\begin{overpic}[width=2.1cm, percent]{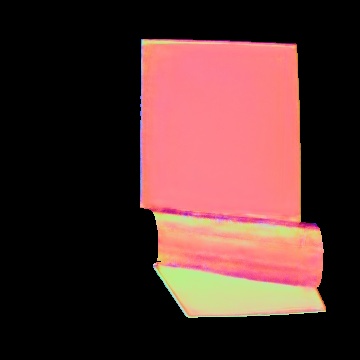}\put(0,85){\color{white} \small 24.02}\end{overpic}\\
			\begin{overpic}[width=2.1cm, percent]{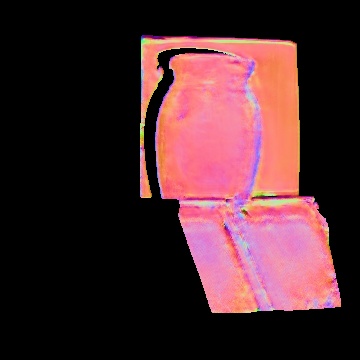}\put(0,85){\color{white} \small 28.09}\end{overpic}\\
			\begin{overpic}[width=2.1cm, percent]{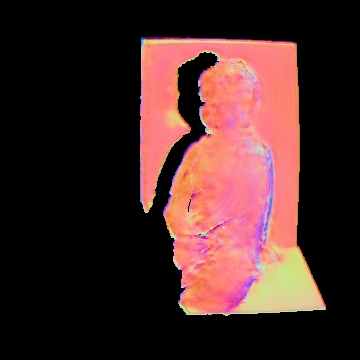}\put(0,85){\color{white} \small 27.00}\end{overpic}\\
			\begin{overpic}[width=2.1cm, percent]{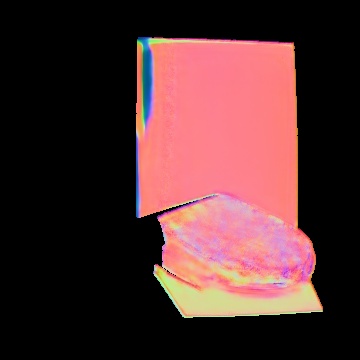}\put(0,85){\color{white} \small 31.50}\end{overpic}\\
			\begin{overpic}[width=2.1cm, percent]{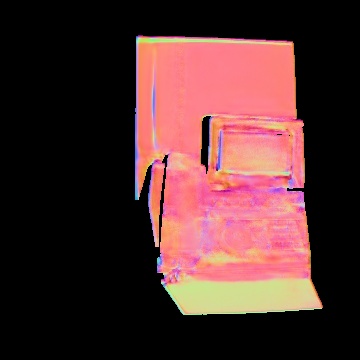}\put(0,85){\color{white} \small 29.42}\end{overpic}\\
			\begin{overpic}[width=2.1cm, percent]{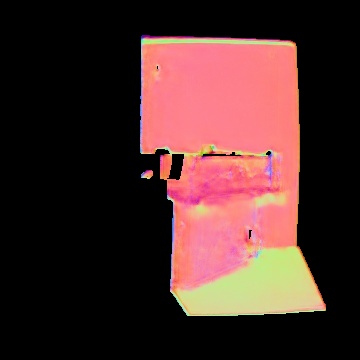}\put(0,85){\color{white} \small 27.66}\end{overpic}\\
			\begin{overpic}[width=2.1cm, percent]{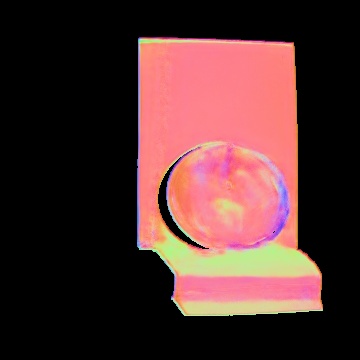}\put(0,85){\color{white} \small 29.49}\end{overpic}
		}{ANNs~\cite{muglikar2023event}\\(b)}
		&
		\column{
			\begin{overpic}[width=2.1cm, percent]{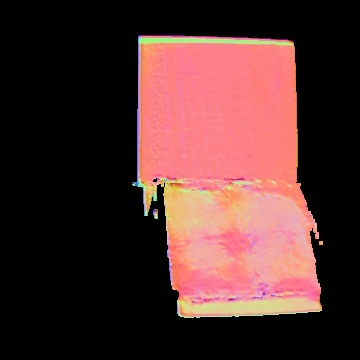}\put(0,85){\color{white} \small 27.77}\end{overpic}\\
			\begin{overpic}[width=2.1cm, percent]{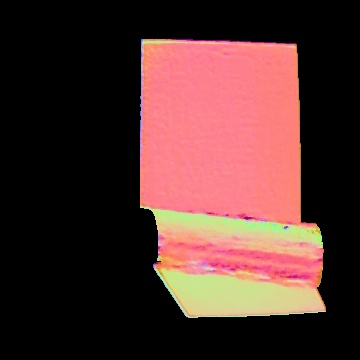}\put(0,85){\color{white} \small 24.11}\end{overpic}\\
			\begin{overpic}[width=2.1cm, percent]{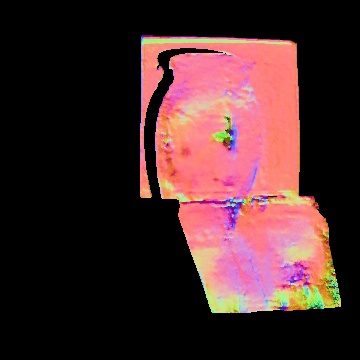}\put(0,85){\color{white} \small 33.17}\end{overpic}\\
			\begin{overpic}[width=2.1cm, percent]{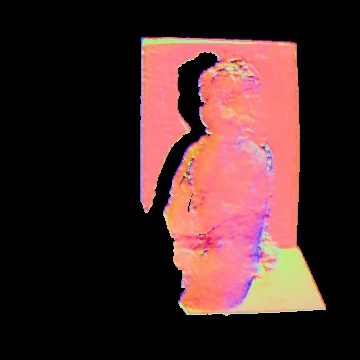}\put(0,85){\color{white} \small 26.84}\end{overpic}\\
			\begin{overpic}[width=2.1cm, percent]{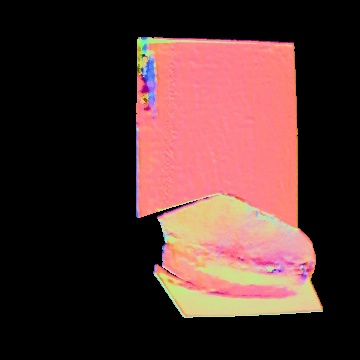}\put(0,85){\color{white} \small 28.94}\end{overpic}\\
			\begin{overpic}[width=2.1cm, percent]{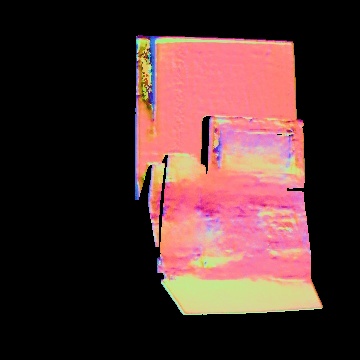}\put(0,85){\color{white} \small 31.41}\end{overpic}\\
			\begin{overpic}[width=2.1cm, percent]{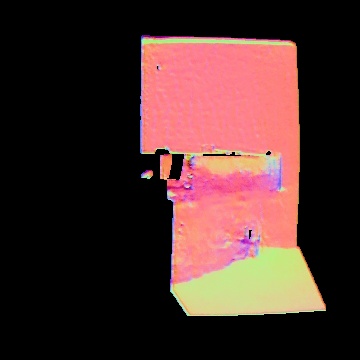}\put(0,85){\color{white} \small 27.73}\end{overpic}\\
			\begin{overpic}[width=2.1cm, percent]{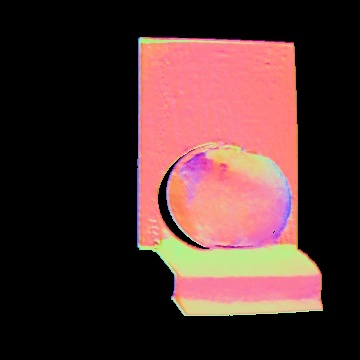}\put(0,85){\color{white} \small 29.57}\end{overpic}
		}{Single\_B\\(c)}
		&
		\column{
			\begin{overpic}[width=2.1cm, percent]{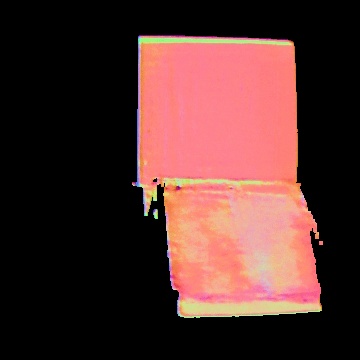}\put(0,85){\color{white} \small 27.65}\end{overpic}\\
			\begin{overpic}[width=2.1cm, percent]{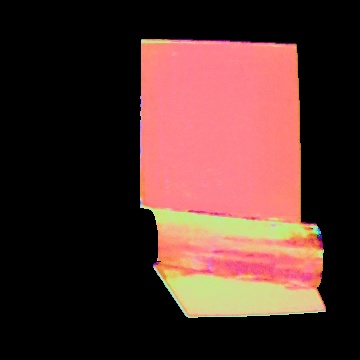}\put(0,85){\color{white} \small 24.12}\end{overpic}\\
			\begin{overpic}[width=2.1cm, percent]{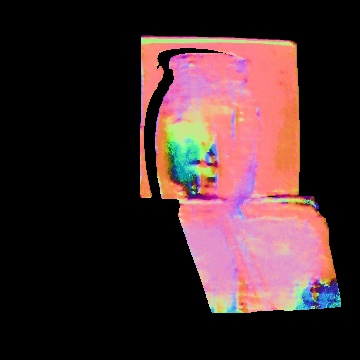}\put(0,85){\color{white} \small 38.20}\end{overpic}\\
			\begin{overpic}[width=2.1cm, percent]{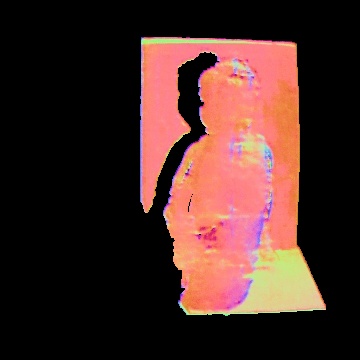}\put(0,85){\color{white} \small 27.46}\end{overpic}\\
			\begin{overpic}[width=2.1cm, percent]{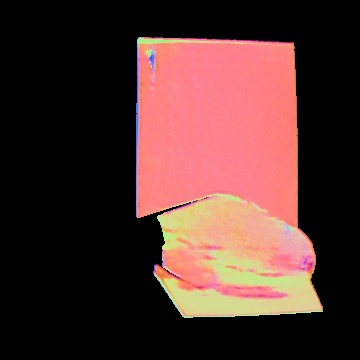}\put(0,85){\color{white} \small 27.88}\end{overpic}\\
			\begin{overpic}[width=2.1cm, percent]{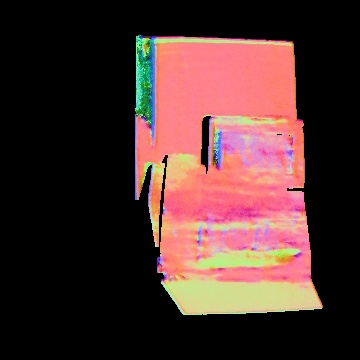}\put(0,85){\color{white} \small 33.94}\end{overpic}\\
			\begin{overpic}[width=2.1cm, percent]{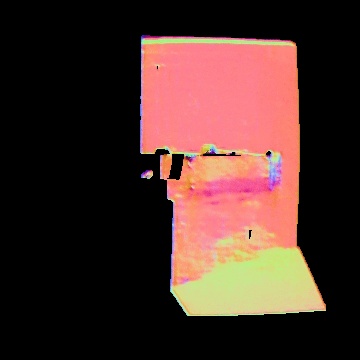}\put(0,85){\color{white} \small 29.07}\end{overpic}\\
			\begin{overpic}[width=2.1cm, percent]{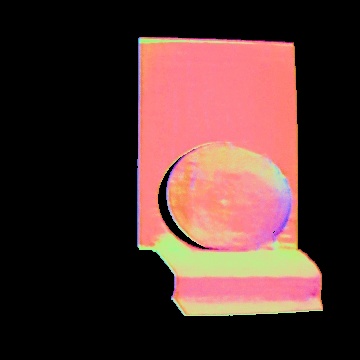}\put(0,85){\color{white} \small 30.03}\end{overpic}
		}{Single\_N\\(d)}
		&
		\column{
			\begin{overpic}[width=2.1cm, percent]{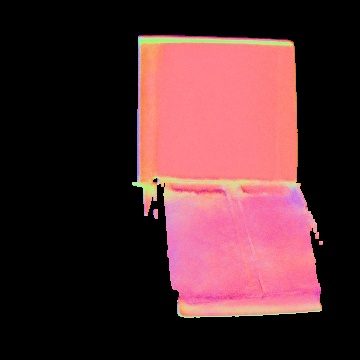}\put(0,85){\color{white} \small 30.27}\end{overpic}\\
			\begin{overpic}[width=2.1cm, percent]{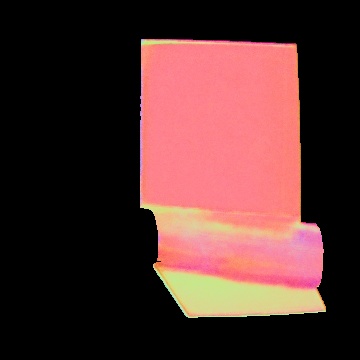}\put(0,85){\color{white} \small 23.76}\end{overpic}\\
			\begin{overpic}[width=2.1cm, percent]{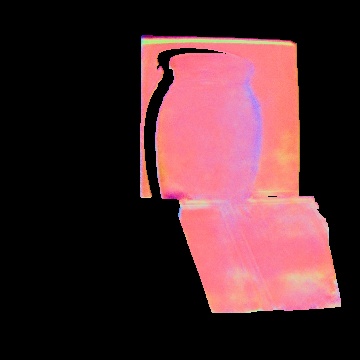}\put(0,85){\color{white} \small 26.79}\end{overpic}\\
			\begin{overpic}[width=2.1cm, percent]{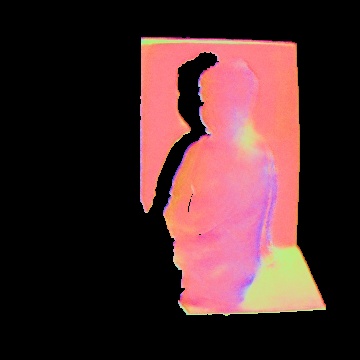}\put(0,85){\color{white} \small 26.72}\end{overpic}\\
			\begin{overpic}[width=2.1cm, percent]{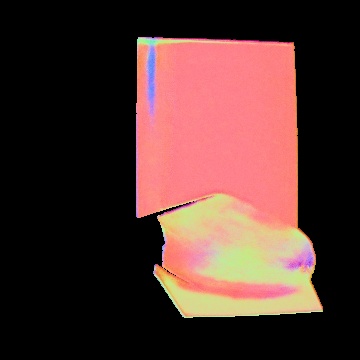}\put(0,85){\color{white} \small 27.71}\end{overpic}\\
			\begin{overpic}[width=2.1cm, percent]{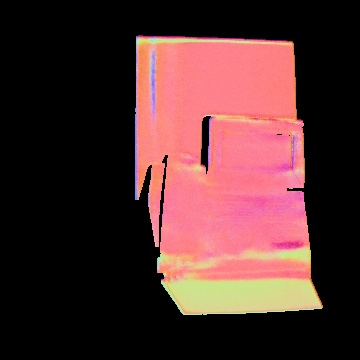}\put(0,85){\color{white} \small 28.76}\end{overpic}\\
			\begin{overpic}[width=2.1cm, percent]{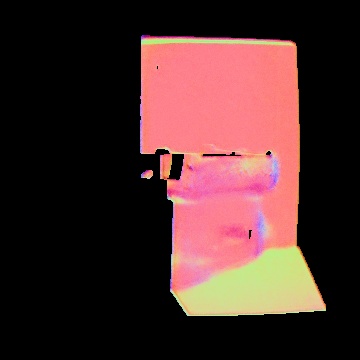}\put(0,85){\colorbox{red}{\color{white} \small 26.05}}\end{overpic}\\
			\begin{overpic}[width=2.1cm, percent]{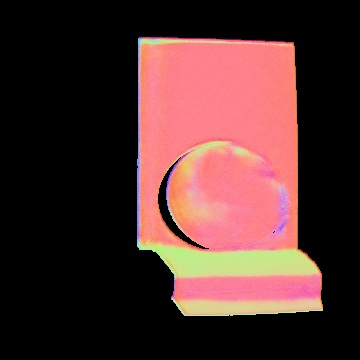}\put(0,85){\colorbox{red}{\color{white} \small 28.60}}\end{overpic}
		}{Multi\_B\\(e)}
		&
		\column{
			\begin{overpic}[width=2.1cm, percent]{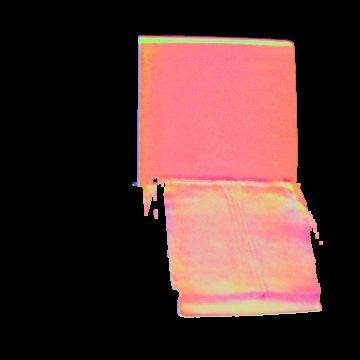}\put(0,85){\colorbox{red}{\color{white} \small 26.67}}\end{overpic}\\
			\begin{overpic}[width=2.1cm, percent]{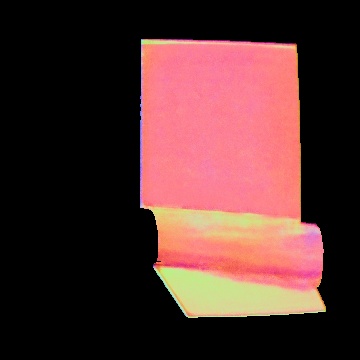}\put(0,85){\colorbox{red}{\color{white} \small 23.36}}\end{overpic}\\
			\begin{overpic}[width=2.1cm, percent]{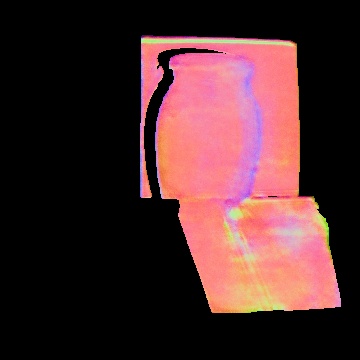}\put(0,85){\colorbox{red}{\color{white} \small 26.65}}\end{overpic}\\
			\begin{overpic}[width=2.1cm, percent]{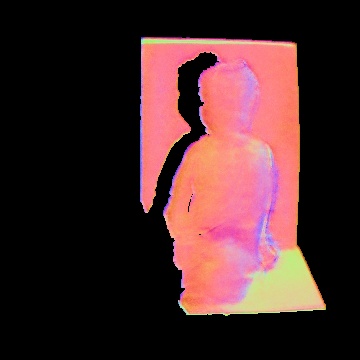}\put(0,85){\colorbox{red}{\color{white} \small 26.19}}\end{overpic}\\
			\begin{overpic}[width=2.1cm, percent]{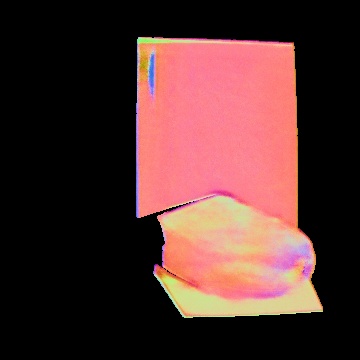}\put(0,85){\colorbox{red}{\color{white} \small 27.13}}\end{overpic}\\
			\begin{overpic}[width=2.1cm, percent]{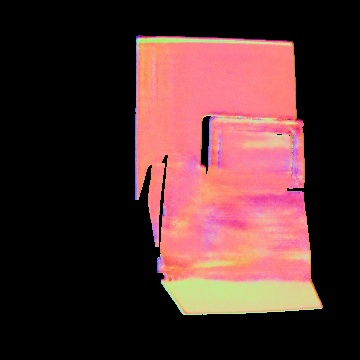}\put(0,85){\colorbox{red}{\color{white} \small 28.73}}\end{overpic}\\
			\begin{overpic}[width=2.1cm, percent]{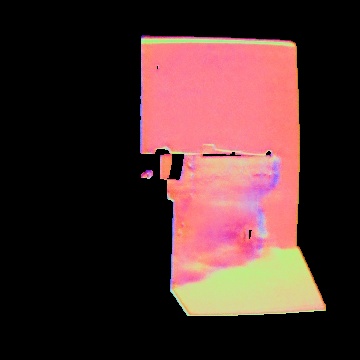}\put(0,85){\color{white} \small 26.35}\end{overpic}\\
			\begin{overpic}[width=2.1cm, percent]{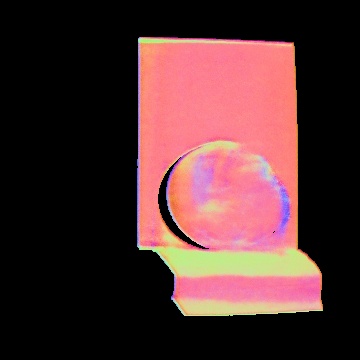}\put(0,85){\color{white} \small 29.34}\end{overpic}
		}{Multi\_N\\(f)}
		&
		\column{
			\begin{overpic}[width=2.1cm, percent]{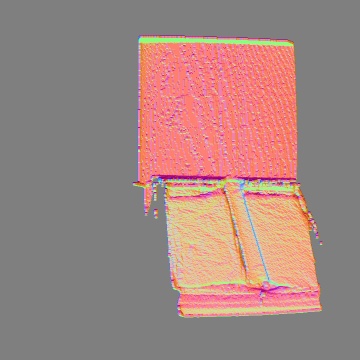}\end{overpic}\\
			\begin{overpic}[width=2.1cm, percent]{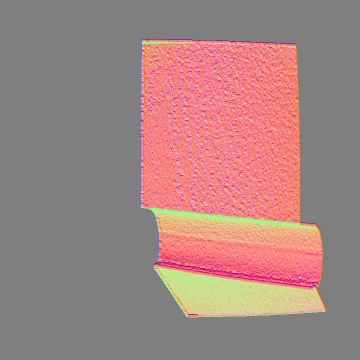}\end{overpic}\\
			\begin{overpic}[width=2.1cm, percent]{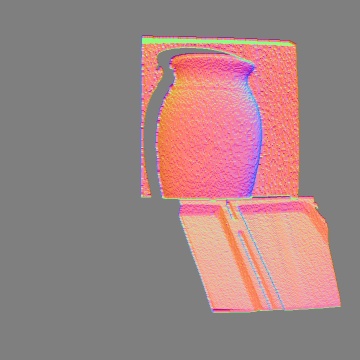}\end{overpic}\\
			\begin{overpic}[width=2.1cm, percent]{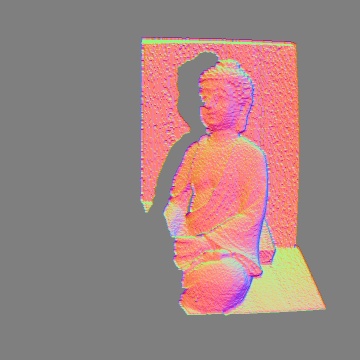}\end{overpic}\\
			\begin{overpic}[width=2.1cm, percent]{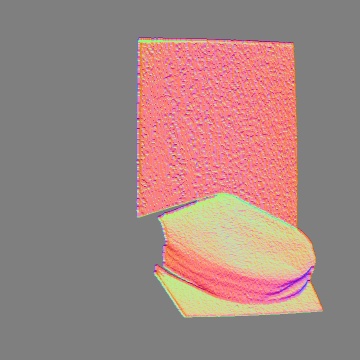}\end{overpic}\\
			\begin{overpic}[width=2.1cm, percent]{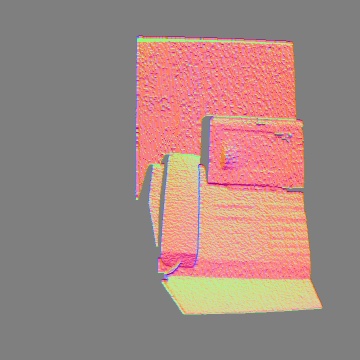}\end{overpic}\\
			\begin{overpic}[width=2.1cm, percent]{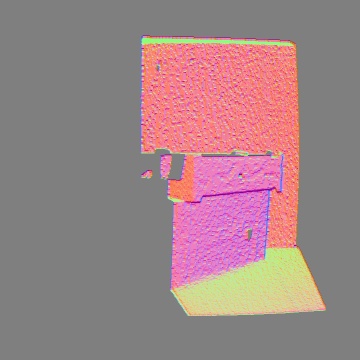}\end{overpic}\\
			\begin{overpic}[width=2.1cm, percent]{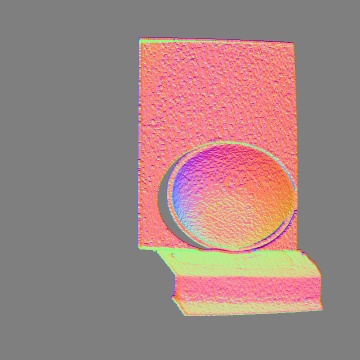}\end{overpic}
		}{GT\\(g)}
	\end{tabular}
	
	\caption{\label{exp2}Qualitative results on the ESfP-Real Dataset. Column (a) shows the scene photographs for context. Column (b) is for the counterpart ANN models. Columns (c-d) are for the Single-Timestep Spiking UNets with bilinear upsampling and nearest neighbor upsampling, respectively. Columns (e-f) are for the Multi-Timestep Spiking UNets with bilinear upsampling and nearest neighbor upsampling, respectively. Column (g) presents the ground truth normals. The MAE for the reconstructions is shown on the top left of each cell in Columns (b-f). For each scene, we highlight the best result using the red colorbox.}
	
\end{figure}

We also compare these methods on the ESfP-Real Dataset. Specifically, we show the quantitative performance in Table~\ref{t3} and illustrate the qualitative results in Fig.~\ref{exp2}. Similar to the results on the ESfP-Synthetic Dataset, our models demonstrate superior performance compared to physics-based methods on the real-world dataset. Moreover, as indicated by Table~\ref{t3} and Fig.~\ref{exp2}, our models not only match the overall performance of the ANN counterpart but also excel in qualitative results across diverse scenes in the test dataset. This enhanced performance on the ESfP-Real Dataset can be attributed to the sparser nature of this real-world dataset~\cite{muglikar2023event}. In addition, compared to the ANN counterpart, our model is more compatible with the sparse events and better maintains the sparsity to prevent overfitting on this dataset. 

Mirroring the outcomes observed on the ESfP-Synthetic Dataset, results from Table~\ref{t3} and Fig.~\ref{exp2} also show that the Multi-Timestep Spiking UNet slightly outperforms the Single-Timestep Spiking UNet. Additionally, nearest neighbor upsampling is on par with bilinear upsampling in terms of surface normal estimation performance.

\begin{table}[]
	\caption{\label{t3}Shape from polarization performance on the ESfP-Real Dataset in terms of Mean Angular Error (MAE) and the percentage of pixels under specific angular errors (AE$<\cdot$). The "Input" column specifies whether the method utilizes events (E) or polarization images (I). E+I[0] means the CVGR-I representation. ``Single" is for the Single-Timestep Spiking UNet. ``Multi'' is for the Multi-Timestep Spiking UNet. ``Bilinear'' and ``Nearest'' represent the bilinear upsampling and nearest neighbor upsampling, respectively. We highlight the top performance in bold, and underline the second-best results.}
	\begin{tabular}{l|c|c|c|c|c|c}
		\hline
		Method           & Input  & Task           & MAE$\downarrow$             & AE\textless{}11.25$\uparrow$ & AE\textless{}22.5$\uparrow$ & AE\textless{}30$\uparrow$ \\ \hline
		Mahmoud \textit{et al.}~\cite{mahmoud2012direct}         & I    & Physics  & 56.278          & 0.032              & 0.091             & 0.163           \\ \hline
		Smith \textit{et al.}~\cite{smith2018height}            & I    & Physics  & 72.525         & 0.009           & 0.034             & 0.058       \\ \hline
		Muglikar \textit{et al.}~\cite{muglikar2023event}          & E    & Physics & 38.786 & 0.087    & 0.220   & 0.452 \\ \hline
		Muglikar \textit{et al.}~\cite{muglikar2023event}         & E+I[0] & Learning & \underline{26.851}
        & 0.099             & 0.449             & \textbf{\small 0.691 }          \\ \hline
		Single\_Bilinear & E+I[0] & Learning & 27.134          & \textbf{\small 0.109}             & \textbf{\small 0.458}             & 0.685           \\ \hline
		Single\_Nearest  & E+I[0] & Learning & 27.391          & \underline{0.106}              & \underline{0.450}             & 0.684           \\ \hline
		Multi\_Bilinear  & E+I[0] & Learning & 26.886          & 0.093              & 0.439             & \underline{0.689}           \\ \hline
		Multi\_Nearest   & E+I[0] & Learning & \textbf{\small 26.781 }   & 0.089       & \underline{0.450}       & 0.688     \\ \hline
		
	\end{tabular}
\end{table}

\subsection{Energy Analysis}

In earlier sections, we demonstrated that our models, employing nearest neighbor upsampling, can achieve performance comparable to those using bilinear upsampling in event-based shape from polarization. To delve deeper into the advantages of these fully spiking models, we will now estimate the computational cost savings they offer compared to their fully ANN counterpart~\cite{muglikar2023event} on the ESfP-Real Dataset. Commonly, the number of synaptic operations serves as a benchmark for assessing the computational energy of SNN models, as referenced in studies like \cite{ijcai2021-441} and \cite{lee2020spike}. Moreover, we can approximate the total energy consumption of a model using principles based on CMOS technology, as outlined in \cite{horowitz20141}.

Unlike ANNs, which consistently perform real-valued matrix-vector multiplication operations regardless of input sparsity, SNNs execute computations based on events, triggered only upon receiving input spikes. Therefore, we initially assess the mean spiking rate of layer $l$ in our proposed model. In particular, the mean spiking rate for layer $l$ in an SNN is calculated as follows:
\begin{equation}
	\label{e-fr0}
	F^{(l)} = \frac{1}{T}\sum_{t\in T}\frac{S^{(l)}_t}{K^{(l)}}
\end{equation}
where $T$ is the total time length, $S^{(l)}_t$ is the number of spikes of layer $l$ at time $t$, and $K^{(l)}$ is the number of neurons of layer $l$. Table~\ref{tab:enery1} shows the mean spiking rates for all layers in our fully spiking models, including the Single-Timestep Spiking UNet and Multi-Timestep Spiking UNet. Notice that we do not consider the components without trainable weights, such as max pooling and nearest neighbor upsampling layers. From the table, we can see that the Multi-Timestep Spiking UNet exhibits a higher average spiking rate across all its layers compared to the Single-Timestep Spiking UNet. This increased spiking rate aids in preserving more information, thereby enhancing the accuracy of surface normal estimation.

With the mean spiking rates, we can estimate the number of synaptic operations in the SNNs. Given $M$ is the number of neurons, $C$ is the number of synaptic connections per neuron, and $F$ indicates the mean spiking rate, the number of synaptic operations at each time in layer $l$ is calculated as $M^{(l)}\times C^{(l)}\times F^{(l)}$. Thus, the total number of synaptic operations in an SNN is calculated by: 
\begin{equation}
	\label{e-fr2}
	\#OP = \sum_{l}M^{(l)}\times C^{(l)}\times F^{(l)}\times T.
\end{equation}
In contrast, the total number of synaptic operations in the ANNs is $\sum_lM^{(l)}\times C^{(l)}$. Due to the binary nature of spikes, SNNs perform only accumulation (AC) per synaptic operation, while ANNs perform the multiply-accumulate (MAC) computations since the operations are real-valued. Based on these, we estimate the number of synaptic operations in the our proposed models and their ANN counterpart. Table~\ref{tab:enery2} illustrates that, in comparison to ANNs, our models primarily perform AC operations with significantly fewer MAC operations that transform real-valued event inputs into binary spiking representations. Furthermore, the Multi-Timestep Spiking UNet executes more AC operations than the Single-Timestep Spiking UNet due to its higher average spiking rate and the utilization of temporal dynamics across multiple timesteps.

In general, AC operation is considered to be significantly more energy-efficient than MAC. For example, an AC is reported to be $\mathbf{5.1\times}$ more energy-efficient than a MAC in the case of 32-bit floating-point numbers (0.9pJ vs. 4.6pJ, 45nm CMOS process)~\cite{horowitz20141}. Based on this principle, we obtain the computational energy benefits of SNNs over ANNs in Table~\ref{tab:enery2}. From the table, we can see that the SNN models are \textbf{3.14$\times$} to \textbf{28.80$\times$} more energy-efficient than ANNs on the ESfP-Real Dataset. 


These results are consistent with the fact that the sparse spike communication and event-driven computation underlie the efficiency advantage of SNNs and demonstrate the potential of our models on neuromorphic hardware and energy-constrained devices.

%
%

\begin{table}[]
	\centering
	\caption{\label{tab:enery1}Mean spiking rates for all layers in the Single-Timestep Spiking UNet and Multi-Timestep Spiking UNet, both utilizing nearest neighbor upsampling and being fully spiking. Layers 1 to 19 correspond to the spiking convolutional layers depicted in Fig.~\ref{single} and Fig.~\ref{multi}. Given that the CVGR-I inputs are real-valued, the first layer in both models does not involve spike calculation.}
	\resizebox{\textwidth}{!}{%
		\begin{tabular}{l|cc|cc}
			\hline
			\multirow{2}{*}{} & \multicolumn{2}{c|}{Single-Timestep Spiking UNet\_Nearest} & \multicolumn{2}{c}{Multi-Timestep Spiking UNet\_Nearest} \\ \cline{2-5} 
			& \multicolumn{1}{c|}{Spiking rates}   & Spikes   & \multicolumn{1}{c|}{Spiking rates}    & Spikes   \\ \hline
			Layer 1  & \multicolumn{1}{c|}{0.3070} & No & \multicolumn{1}{c|}{0.3070} & No  \\ \hline
			Layer 2  & \multicolumn{1}{c|}{0.0901} & Yes & \multicolumn{1}{c|}{0.2484} & Yes  \\ \hline
			Layer 3  & \multicolumn{1}{c|}{0.1342} & Yes & \multicolumn{1}{c|}{0.2304} & Yes  \\ \hline
			Layer 4  & \multicolumn{1}{c|}{0.1057} & Yes & \multicolumn{1}{c|}{0.1626} & Yes  \\ \hline
			Layer 5  & \multicolumn{1}{c|}{0.1467} & Yes & \multicolumn{1}{c|}{0.2482} & Yes  \\ \hline
			Layer 6  & \multicolumn{1}{c|}{0.1174} & Yes & \multicolumn{1}{c|}{0.1719} & Yes \\ \hline
			Layer 7  & \multicolumn{1}{c|}{0.1485} & Yes & \multicolumn{1}{c|}{0.2733} & Yes  \\ \hline
			Layer 8  & \multicolumn{1}{c|}{0.1153} & Yes & \multicolumn{1}{c|}{0.1870} & Yes  \\ \hline
			Layer 9  & \multicolumn{1}{c|}{0.1717} & Yes & \multicolumn{1}{c|}{0.3607} & Yes  \\ \hline
			Layer 10 & \multicolumn{1}{c|}{0.1691} & Yes  & \multicolumn{1}{c|}{0.2149} & Yes  \\ \hline
			Layer 11 & \multicolumn{1}{c|}{0.1278} & Yes & \multicolumn{1}{c|}{0.1991} & Yes   \\ \hline
			Layer 12 & \multicolumn{1}{c|}{0.1513} & Yes & \multicolumn{1}{c|}{0.2075} & Yes  \\ \hline
			Layer 13 & \multicolumn{1}{c|}{0.1175} & Yes  & \multicolumn{1}{c|}{0.1840} & Yes   \\ \hline
			Layer 14 & \multicolumn{1}{c|}{0.1540} & Yes  & \multicolumn{1}{c|}{0.1923} & Yes \\ \hline
			Layer 15 & \multicolumn{1}{c|}{0.1391} & Yes  & \multicolumn{1}{c|}{0.1810} & Yes   \\ \hline
			Layer 16 & \multicolumn{1}{c|}{0.1937} & Yes  & \multicolumn{1}{c|}{0.1867} & Yes  \\ \hline
			Layer 17 & \multicolumn{1}{c|}{0.1624} & Yes  & \multicolumn{1}{c|}{0.1881} & Yes   \\ \hline
			Layer 18 & \multicolumn{1}{c|}{0.2323} & Yes & \multicolumn{1}{c|}{0.2058} & Yes  \\ \hline
			Layer 19 & \multicolumn{1}{c|}{0.2080} & Yes  & \multicolumn{1}{c|}{0.2099} & Yes  \\ \hline
			Average & \multicolumn{1}{c|}{0.1575} & - & \multicolumn{1}{c|}{0.2189} & - \\ \hline
		\end{tabular}%
	}
\end{table}

\begin{table}[]
	\centering
	\caption{\label{tab:enery2}Energy comparison of our models and their ANN counterpart on the ESfP-Real Dataset. The energy benefit is equal to $Energy_{ANNs} / Energy_{SNNs}$.}
	\begin{tabular}{l|c|c|c}
		\hline
		& ANNs~\cite{muglikar2023event} & Single\_Nearest & Multi\_Nearest \\ \hline
		Average Spiking Rate                 &           -                           &                0.1575                       &         0.2189                             \\ \hline
		\#OP\_{MAC} ($\times 10^9$)                        &        161.11                                &          1.21                             &         1.21                               \\ \hline
		\#OP\_AC ($\times 10^9$)                          &                   0                     &          22.36                             &        255.35                              \\ \hline
		Energy ($10^{-3}$J, 45nm CMOS process) &          741.11                                &             25.69                        &               235.38                       \\ \hline
		Energy Benefit ($\times$)                      &                 1.0                       &                 28.80                      &              3.14                        \\ \hline
	\end{tabular}
\end{table}

\section{Conclusion and Future Work}

In this work, we explore the domain of event-based shape from polarization with SNNs. Drawing inspiration from the feed-forward UNet, we introduce the Single-Timestep Spiking UNet, which processes event-based shape from polarization as a non-temporal task, updating the membrane potential of each spiking neuron only once. This method, while not fully leveraging the temporal capabilities of SNNs, significantly cuts down on computational and energy demands. To better harness the rich temporal data in event-based information, we also propose the Multi-Timestep Spiking UNet. This model operates sequentially across multiple timesteps, enabling spiking neurons to employ their temporal recurrent neuronal dynamics for more effective data extraction. Through extensive evaluation on both synthetic and real-world datasets, our models demonstrate their ability to estimate dense surface normals from polarization events, achieving results comparable to those of state-of-the-art ANN models. Moreover, our models present enhanced energy efficiency over their ANN counterparts, underscoring their suitability for neuromorphic hardware and energy-sensitive edge devices. This research not only advances the field of spiking neural networks but also opens up new possibilities for efficient and effective event-based shape recovery in various applications.

Building on this foundation, future work could focus on several promising directions. One key area is the further optimization of SNN architectures to enhance their ability to process complex, dynamic scenes, potentially by integrating more sophisticated temporal dynamics or learning algorithms. Additionally, exploring the integration of our models with other sensory data types, like depth information, could lead to more robust and versatile systems. Moreover, adapting these models for real-time applications in various fields, from autonomous vehicles to augmented reality, presents an exciting challenge. Finally, there is significant potential in further reducing the energy consumption of these networks, making them even more suitable for deployment in low-power, edge computing scenarios. Through these explorations, we can continue to push the boundaries of what's possible with SNNs in event-based sensing and beyond.

\section*{Acknowledgement}
We are grateful to Chenghong Lin for her proofreading and advice on the paper writing. 

\section*{References}
\bibliographystyle{unsrt} 
\bibliography{test}

\begin{thebibliography}{10}

\bibitem{kazhdan2006poisson}
Michael Kazhdan, Matthew Bolitho, and Hugues Hoppe.
\newblock Poisson surface reconstruction.
\newblock In {\em Proceedings of the fourth Eurographics symposium on Geometry
  processing}, volume~7, page~0, 2006.

\bibitem{luo2020niid}
Jundan Luo, Zhaoyang Huang, Yijin Li, Xiaowei Zhou, Guofeng Zhang, and Hujun
  Bao.
\newblock Niid-net: adapting surface normal knowledge for intrinsic image
  decomposition in indoor scenes.
\newblock {\em IEEE Transactions on Visualization and Computer Graphics},
  26(12):3434--3445, 2020.

\bibitem{ju2021recovering}
Yakun Ju, Junyu Dong, and Sheng Chen.
\newblock Recovering surface normal and arbitrary images: A dual regression
  network for photometric stereo.
\newblock {\em IEEE Transactions on Image Processing}, 30:3676--3690, 2021.

\bibitem{strese2016multimodal}
Matti Strese, Clemens Schuwerk, Albert Iepure, and Eckehard Steinbach.
\newblock Multimodal feature-based surface material classification.
\newblock {\em IEEE transactions on haptics}, 10(2):226--239, 2016.

\bibitem{badino2011fast}
Hernan Badino, Daniel Huber, Yongwoon Park, and Takeo Kanade.
\newblock Fast and accurate computation of surface normals from range images.
\newblock In {\em 2011 IEEE International Conference on Robotics and
  Automation}, pages 3084--3091. IEEE, 2011.

\bibitem{geng2011structured}
Jason Geng.
\newblock Structured-light 3d surface imaging: a tutorial.
\newblock {\em Advances in optics and photonics}, 3(2):128--160, 2011.

\bibitem{horn1970shape}
Berthold~KP Horn.
\newblock Shape from shading: A method for obtaining the shape of a smooth
  opaque object from one view.
\newblock 1970.

\bibitem{shafer1983using}
Steven~A Shafer and Takeo Kanade.
\newblock Using shadows in finding surface orientations.
\newblock {\em Computer Vision, Graphics, and Image Processing},
  22(1):145--176, 1983.

\bibitem{witkin1981recovering}
Andrew~P Witkin.
\newblock Recovering surface shape and orientation from texture.
\newblock {\em Artificial intelligence}, 17(1-3):17--45, 1981.

\bibitem{hernandez2004stereo}
Carlos Hern{\'a}ndez.
\newblock Stereo and silhouette fusion for 3d object modeling from uncalibrated
  images under circular motion.
\newblock {\em These de doctorat, {\'E}cole Nationale Sup{\'e}rieure des
  T{\'e}l{\'e}communications}, 2, 2004.

\bibitem{woodham1980photometric}
Robert~J Woodham.
\newblock Photometric method for determining surface orientation from multiple
  images.
\newblock {\em Optical engineering}, 19(1):139--144, 1980.

\bibitem{rahmann2001reconstruction}
Stefan Rahmann and Nikos Canterakis.
\newblock Reconstruction of specular surfaces using polarization imaging.
\newblock In {\em Proceedings of the 2001 IEEE Computer Society Conference on
  Computer Vision and Pattern Recognition. CVPR 2001}, volume~1, pages I--I.
  IEEE, 2001.

\bibitem{kadambi2015polarized}
Achuta Kadambi, Vage Taamazyan, Boxin Shi, and Ramesh Raskar.
\newblock Polarized 3d: High-quality depth sensing with polarization cues.
\newblock In {\em Proceedings of the IEEE International Conference on Computer
  Vision}, pages 3370--3378, 2015.

\bibitem{wolff1993constraining}
Lawrence~B Wolff and Terrance~E Boult.
\newblock Constraining object features using a polarization reflectance model.
\newblock {\em Phys. Based Vis. Princ. Pract. Radiom}, 1:167, 1993.

\bibitem{tozza2017linear}
Silvia Tozza, William~AP Smith, Dizhong Zhu, Ravi Ramamoorthi, and Edwin~R
  Hancock.
\newblock Linear differential constraints for photo-polarimetric height
  estimation.
\newblock In {\em Proceedings of the IEEE international conference on computer
  vision}, pages 2279--2287, 2017.

\bibitem{ba2020deep}
Yunhao Ba, Alex Gilbert, Franklin Wang, Jinfa Yang, Rui Chen, Yiqin Wang, Lei
  Yan, Boxin Shi, and Achuta Kadambi.
\newblock Deep shape from polarization.
\newblock In {\em Computer Vision--ECCV 2020: 16th European Conference,
  Glasgow, UK, August 23--28, 2020, Proceedings, Part XXIV 16}, pages 554--571.
  Springer, 2020.

\bibitem{lei2022shape}
Chenyang Lei, Chenyang Qi, Jiaxin Xie, Na~Fan, Vladlen Koltun, and Qifeng Chen.
\newblock Shape from polarization for complex scenes in the wild.
\newblock In {\em Proceedings of the IEEE/CVF Conference on Computer Vision and
  Pattern Recognition}, pages 12632--12641, 2022.

\bibitem{kadambi2017depth}
Achuta Kadambi, Vage Taamazyan, Boxin Shi, and Ramesh Raskar.
\newblock Depth sensing using geometrically constrained polarization normals.
\newblock {\em International Journal of Computer Vision}, 125:34--51, 2017.

\bibitem{atkinson2018high}
Gary~A Atkinson and J{\"u}rgen~D Ernst.
\newblock High-sensitivity analysis of polarization by surface reflection.
\newblock {\em Machine Vision and Applications}, 29:1171--1189, 2018.

\bibitem{wolff1997polarization}
Lawrence~B Wolff.
\newblock Polarization vision: a new sensory approach to image understanding.
\newblock {\em Image and Vision computing}, 15(2):81--93, 1997.

\bibitem{LucidV}
Lucid vision phoenix polarization camera.
\newblock \url{https://thinklucid.com/product/phoenix-5-0-mp-polarized-model/}.
\newblock 2018.

\bibitem{muglikar2023event}
Manasi Muglikar, Leonard Bauersfeld, Diederik~Paul Moeys, and Davide
  Scaramuzza.
\newblock Event-based shape from polarization.
\newblock In {\em Proceedings of the IEEE/CVF Conference on Computer Vision and
  Pattern Recognition}, pages 1547--1556, 2023.

\bibitem{gallego2020event}
Guillermo Gallego, Tobi Delbruck, Garrick~Michael Orchard, Chiara Bartolozzi,
  Brian Taba, Andrea Censi, Stefan Leutenegger, Andrew Davison, Jorg Conradt,
  Kostas Daniilidis, et~al.
\newblock Event-based vision: A survey.
\newblock {\em IEEE transactions on pattern analysis and machine intelligence},
  2020.

\bibitem{roy2019towards}
Kaushik Roy, Akhilesh Jaiswal, and Priyadarshini Panda.
\newblock Towards spike-based machine intelligence with neuromorphic computing.
\newblock {\em Nature}, 575(7784):607--617, 2019.

\bibitem{maas2013rectifier}
Andrew~L Maas, Awni~Y Hannun, Andrew~Y Ng, et~al.
\newblock Rectifier nonlinearities improve neural network acoustic models.
\newblock In {\em Proc. icml}, volume~30, page~3. Citeseer, 2013.

\bibitem{xu2015empirical}
Bing Xu, Naiyan Wang, Tianqi Chen, and Mu~Li.
\newblock Empirical evaluation of rectified activations in convolutional
  network.
\newblock {\em arXiv preprint arXiv:1505.00853}, 2015.

\bibitem{clevert2015fast}
Djork-Arn{\'e} Clevert, Thomas Unterthiner, and Sepp Hochreiter.
\newblock Fast and accurate deep network learning by exponential linear units
  (elus).
\newblock {\em arXiv preprint arXiv:1511.07289}, 2015.

\bibitem{gerstner1995time}
Wulfram Gerstner.
\newblock Time structure of the activity in neural network models.
\newblock {\em Physical review E}, 51(1):738, 1995.

\bibitem{abbott1999lapicque}
Larry~F Abbott.
\newblock Lapicque’s introduction of the integrate-and-fire model neuron
  (1907).
\newblock {\em Brain research bulletin}, 50(5-6):303--304, 1999.

\bibitem{gerstner2002spiking}
Wulfram Gerstner and Werner~M Kistler.
\newblock {\em Spiking neuron models: Single neurons, populations, plasticity}.
\newblock Cambridge university press, 2002.

\bibitem{lecun2015deep}
Yann LeCun, Yoshua Bengio, and Geoffrey Hinton.
\newblock Deep learning.
\newblock {\em nature}, 521(7553):436--444, 2015.

\bibitem{zhou2022spikformer}
Zhaokun Zhou, Yuesheng Zhu, Chao He, Yaowei Wang, Shuicheng Yan, Yonghong Tian,
  and Li~Yuan.
\newblock Spikformer: When spiking neural network meets transformer.
\newblock {\em arXiv preprint arXiv:2209.15425}, 2022.

\bibitem{zhang2022spiking}
Jiqing Zhang, Bo~Dong, Haiwei Zhang, Jianchuan Ding, Felix Heide, Baocai Yin,
  and Xin Yang.
\newblock Spiking transformers for event-based single object tracking.
\newblock In {\em Proceedings of the IEEE/CVF conference on Computer Vision and
  Pattern Recognition}, pages 8801--8810, 2022.

\bibitem{zhu2022spiking}
Zulun Zhu, Jiaying Peng, Jintang Li, Liang Chen, Qi~Yu, and Siqiang Luo.
\newblock Spiking graph convolutional networks.
\newblock {\em arXiv preprint arXiv:2205.02767}, 2022.

\bibitem{zhu2023spikegpt}
Rui-Jie Zhu, Qihang Zhao, and Jason~K Eshraghian.
\newblock Spikegpt: Generative pre-trained language model with spiking neural
  networks.
\newblock {\em arXiv preprint arXiv:2302.13939}, 2023.

\bibitem{ronneberger2015u}
Olaf Ronneberger, Philipp Fischer, and Thomas Brox.
\newblock U-net: Convolutional networks for biomedical image segmentation.
\newblock In {\em Medical Image Computing and Computer-Assisted
  Intervention--MICCAI 2015: 18th International Conference, Munich, Germany,
  October 5-9, 2015, Proceedings, Part III 18}, pages 234--241. Springer, 2015.

\bibitem{shurcliff1962polarized}
William~A Shurcliff.
\newblock {\em Polarized light: production and use}.
\newblock Harvard University Press, 1962.

\bibitem{collett2005field}
Edward Collett.
\newblock Field guide to polarization.
\newblock Spie Bellingham, WA, 2005.

\bibitem{shi2020recent}
Boxin Shi, Jinfa Yang, Jinwei Chen, Ruihua Zhang, and Rui Chen.
\newblock Recent progress in shape from polarization.
\newblock {\em Advances in Photometric 3D-Reconstruction}, pages 177--203,
  2020.

\bibitem{miyazaki2003polarization}
Miyazaki, Kagesawa, and Ikeuchi.
\newblock Polarization-based transparent surface modeling from two views.
\newblock In {\em Proceedings Ninth IEEE International Conference on Computer
  Vision}, pages 1381--1386. IEEE, 2003.

\bibitem{miyazaki2017surface}
Daisuke Miyazaki, Takuya Shigetomi, Masashi Baba, Ryo Furukawa, Shinsaku Hiura,
  and Naoki Asada.
\newblock Surface normal estimation of black specular objects from multiview
  polarization images.
\newblock {\em Optical Engineering}, 56(4):041303--041303, 2017.

\bibitem{yang2018polarimetric}
Luwei Yang, Feitong Tan, Ao~Li, Zhaopeng Cui, Yasutaka Furukawa, and Ping Tan.
\newblock Polarimetric dense monocular slam.
\newblock In {\em Proceedings of the IEEE conference on computer vision and
  pattern recognition}, pages 3857--3866, 2018.

\bibitem{zhu2019depth}
Dizhong Zhu and William~AP Smith.
\newblock Depth from a polarisation+ rgb stereo pair.
\newblock In {\em Proceedings of the IEEE/CVF Conference on Computer Vision and
  Pattern Recognition}, pages 7586--7595, 2019.

\bibitem{stolz2012shape}
Christophe Stolz, Mathias Ferraton, and Fabrice Meriaudeau.
\newblock Shape from polarization: a method for solving zenithal angle
  ambiguity.
\newblock {\em Optics letters}, 37(20):4218--4220, 2012.

\bibitem{wolff1998improved}
Lawrence~B Wolff, Shree~K Nayar, and Michael Oren.
\newblock Improved diffuse reflection models for computer vision.
\newblock {\em International Journal of Computer Vision}, 30:55--71, 1998.

\bibitem{yu2017shape}
Ye~Yu, Dizhong Zhu, and William~AP Smith.
\newblock Shape-from-polarisation: a nonlinear least squares approach.
\newblock In {\em Proceedings of the IEEE International Conference on Computer
  Vision Workshops}, pages 2969--2976, 2017.

\bibitem{smith2018height}
William~AP Smith, Ravi Ramamoorthi, and Silvia Tozza.
\newblock Height-from-polarisation with unknown lighting or albedo.
\newblock {\em IEEE transactions on pattern analysis and machine intelligence},
  41(12):2875--2888, 2018.

\bibitem{marshall1988unique}
N~Justin Marshall.
\newblock A unique colour and polarization vision system in mantis shrimps.
\newblock {\em Nature}, 333(6173):557--560, 1988.

\bibitem{haessig2023pdavis}
Germain Haessig, Damien Joubert, Justin Haque, Moritz~B Milde, Tobi Delbruck,
  and Viktor Gruev.
\newblock Pdavis: Bio-inspired polarization event camera.
\newblock In {\em Proceedings of the IEEE/CVF Conference on Computer Vision and
  Pattern Recognition}, pages 3962--3971, 2023.

\bibitem{zhu2019unsupervised}
Alex~Zihao Zhu, Liangzhe Yuan, Kenneth Chaney, and Kostas Daniilidis.
\newblock Unsupervised event-based learning of optical flow, depth, and
  egomotion.
\newblock In {\em Proceedings of the IEEE/CVF Conference on Computer Vision and
  Pattern Recognition}, pages 989--997, 2019.

\bibitem{bullmore2012economy}
Ed~Bullmore and Olaf Sporns.
\newblock The economy of brain network organization.
\newblock {\em Nature Reviews Neuroscience}, 13(5):336--349, 2012.

\bibitem{felleman1991distributed}
Daniel~J Felleman and David~C Van~Essen.
\newblock Distributed hierarchical processing in the primate cerebral cortex.
\newblock In {\em Cereb cortex}. Citeseer, 1991.

\bibitem{merolla2014million}
Paul~A Merolla, John~V Arthur, Rodrigo Alvarez-Icaza, Andrew~S Cassidy, Jun
  Sawada, Filipp Akopyan, Bryan~L Jackson, Nabil Imam, Chen Guo, Yutaka
  Nakamura, et~al.
\newblock A million spiking-neuron integrated circuit with a scalable
  communication network and interface.
\newblock {\em Science}, 345(6197):668--673, 2014.

\bibitem{davies2021advancing}
Mike Davies, Andreas Wild, Garrick Orchard, Yulia Sandamirskaya, Gabriel
  A~Fonseca Guerra, Prasad Joshi, Philipp Plank, and Sumedh~R Risbud.
\newblock Advancing neuromorphic computing with loihi: A survey of results and
  outlook.
\newblock {\em Proceedings of the IEEE}, 109(5):911--934, 2021.

\bibitem{cao2015spiking}
Yongqiang Cao, Yang Chen, and Deepak Khosla.
\newblock Spiking deep convolutional neural networks for energy-efficient
  object recognition.
\newblock {\em International Journal of Computer Vision}, 113(1):54--66, 2015.

\bibitem{sengupta2019going}
Abhronil Sengupta, Yuting Ye, Robert Wang, Chiao Liu, and Kaushik Roy.
\newblock Going deeper in spiking neural networks: Vgg and residual
  architectures.
\newblock {\em Frontiers in neuroscience}, 13:95, 2019.

\bibitem{wu2018spatio}
Yujie Wu, Lei Deng, Guoqi Li, Jun Zhu, and Luping Shi.
\newblock Spatio-temporal backpropagation for training high-performance spiking
  neural networks.
\newblock {\em Frontiers in neuroscience}, 12:331, 2018.

\bibitem{ijcai2020-211}
Xiang Cheng, Yunzhe Hao, Jiaming Xu, and Bo~Xu.
\newblock Lisnn: Improving spiking neural networks with lateral interactions
  for robust object recognition.
\newblock In Christian Bessiere, editor, {\em Proceedings of the Twenty-Ninth
  International Joint Conference on Artificial Intelligence, {IJCAI-20}}, pages
  1519--1525. International Joint Conferences on Artificial Intelligence
  Organization, 7 2020.
\newblock Main track.

\bibitem{ranccon2022stereospike}
Ulysse Ran{\c{c}}on, Javier Cuadrado-Anibarro, Benoit~R Cottereau, and
  Timoth{\'e}e Masquelier.
\newblock Stereospike: Depth learning with a spiking neural network.
\newblock {\em IEEE Access}, 10:127428--127439, 2022.

\bibitem{wu2022mss}
Xiaoshan Wu, Weihua He, Man Yao, Ziyang Zhang, Yaoyuan Wang, and Guoqi Li.
\newblock Mss-depthnet: Depth prediction with multi-step spiking neural
  network.
\newblock {\em arXiv preprint arXiv:2211.12156}, 2022.

\bibitem{yao2023spiking}
Xingting Yao, Qinghao Hu, Tielong Liu, Zitao Mo, Zeyu Zhu, Zhengyang Zhuge, and
  Jian Cheng.
\newblock Spiking nerf: Making bio-inspired neural networks see through the
  real world.
\newblock {\em arXiv preprint arXiv:2309.10987}, 2023.

\bibitem{rebecq2019high}
Henri Rebecq, Ren{\'e} Ranftl, Vladlen Koltun, and Davide Scaramuzza.
\newblock High speed and high dynamic range video with an event camera.
\newblock {\em IEEE transactions on pattern analysis and machine intelligence},
  43(6):1964--1980, 2019.

\bibitem{tulyakov2022time}
Stepan Tulyakov, Alfredo Bochicchio, Daniel Gehrig, Stamatios Georgoulis,
  Yuanyou Li, and Davide Scaramuzza.
\newblock Time lens++: Event-based frame interpolation with parametric
  non-linear flow and multi-scale fusion.
\newblock In {\em Proceedings of the IEEE/CVF Conference on Computer Vision and
  Pattern Recognition}, pages 17755--17764, 2022.

\bibitem{fang2021incorporating}
Wei Fang, Zhaofei Yu, Yanqi Chen, Timoth{\'e}e Masquelier, Tiejun Huang, and
  Yonghong Tian.
\newblock Incorporating learnable membrane time constant to enhance learning of
  spiking neural networks.
\newblock In {\em Proceedings of the IEEE/CVF international conference on
  computer vision}, pages 2661--2671, 2021.

\bibitem{odena2016deconvolution}
Augustus Odena, Vincent Dumoulin, and Chris Olah.
\newblock Deconvolution and checkerboard artifacts.
\newblock {\em Distill}, 1(10):e3, 2016.

\bibitem{strohmer2021integrating}
Beck Strohmer, Rasmus~Karn{\o}e Stagsted, Poramate Manoonpong, and Leon~Bonde
  Larsen.
\newblock Integrating non-spiking interneurons in spiking neural networks.
\newblock {\em Frontiers in neuroscience}, 15:633945, 2021.

\bibitem{wu2021liaf}
Zhenzhi Wu, Hehui Zhang, Yihan Lin, Guoqi Li, Meng Wang, and Ye~Tang.
\newblock Liaf-net: Leaky integrate and analog fire network for lightweight and
  efficient spatiotemporal information processing.
\newblock {\em IEEE Transactions on Neural Networks and Learning Systems},
  33(11):6249--6262, 2021.

\bibitem{hecht1992theory}
Robert Hecht-Nielsen.
\newblock Theory of the backpropagation neural network.
\newblock In {\em Neural networks for perception}, pages 65--93. Elsevier,
  1992.

\bibitem{werbos1990backpropagation}
Paul~J Werbos.
\newblock Backpropagation through time: what it does and how to do it.
\newblock {\em Proceedings of the IEEE}, 78(10):1550--1560, 1990.

\bibitem{neftci2019surrogate}
Emre~O Neftci, Hesham Mostafa, and Friedemann Zenke.
\newblock Surrogate gradient learning in spiking neural networks: Bringing the
  power of gradient-based optimization to spiking neural networks.
\newblock {\em IEEE Signal Processing Magazine}, 36(6):51--63, 2019.

\bibitem{fang2023spikingjelly}
Wei Fang, Yanqi Chen, Jianhao Ding, Zhaofei Yu, Timoth{\'e}e Masquelier, Ding
  Chen, Liwei Huang, Huihui Zhou, Guoqi Li, and Yonghong Tian.
\newblock Spikingjelly: An open-source machine learning infrastructure platform
  for spike-based intelligence.
\newblock {\em Science Advances}, 9(40):eadi1480, 2023.

\bibitem{paszke2019pytorch}
Adam Paszke, Sam Gross, Francisco Massa, Adam Lerer, James Bradbury, Gregory
  Chanan, Trevor Killeen, Zeming Lin, Natalia Gimelshein, Luca Antiga, et~al.
\newblock Pytorch: An imperative style, high-performance deep learning library.
\newblock {\em Advances in neural information processing systems}, 32, 2019.

\bibitem{ledinauskas2020training}
Eimantas Ledinauskas, Julius Ruseckas, Alfonsas Jur{\v{s}}{\.e}nas, and
  Giedrius Bura{\v{c}}as.
\newblock Training deep spiking neural networks.
\newblock {\em arXiv preprint arXiv:2006.04436}, 2020.

\bibitem{kingma2014adam}
Diederik~P Kingma and Jimmy Ba.
\newblock Adam: A method for stochastic optimization.
\newblock {\em arXiv preprint arXiv:1412.6980}, 2014.

\bibitem{Mitsuba3}
Wenzel Jakob, Sébastien Speierer, Nicolas Roussel, Merlin Nimier-David, Delio
  Vicini, Tizian Zeltner, Baptiste Nicolet, Miguel Crespo, Vincent Leroy, and
  Ziyi Zhang.
\newblock Mitsuba 3 renderer, 2022.
\newblock https://mitsuba-renderer.org.

\bibitem{rebecq2018esim}
Henri Rebecq, Daniel Gehrig, and Davide Scaramuzza.
\newblock Esim: an open event camera simulator.
\newblock In {\em Conference on robot learning}, pages 969--982. PMLR, 2018.

\bibitem{finateu20201280x720}
T~Finateu, A~Niwa, D~Matolin, K~Tsuchimoto, A~Mascheroni, E~Reynaud,
  P~Mostafalu, F~Brady, L~Chotard, F~LeGoff, et~al.
\newblock A 1280x720 back-illuminated stacked temporal contrast event-based
  vision sensor with 4.86 um pixels, 1.066 geps readout, programmable
  event-rate controller and compressive data-formatting pipeline.
\newblock In {\em IEEE International Solid-State Circuits Conference}, 2020.

\bibitem{X4CPL}
Breakthrough photography x4 polarizer.
\newblock
  \url{https://breakthrough.photography/products/x4-circular-polarizer}.

\bibitem{muglikar2021esl}
Manasi Muglikar, Guillermo Gallego, and Davide Scaramuzza.
\newblock Esl: Event-based structured light.
\newblock In {\em 2021 International Conference on 3D Vision (3DV)}, pages
  1165--1174. IEEE, 2021.

\bibitem{mahmoud2012direct}
Ali~H Mahmoud, Moumen~T El-Melegy, and Aly~A Farag.
\newblock Direct method for shape recovery from polarization and shading.
\newblock In {\em 2012 19th IEEE International Conference on Image Processing},
  pages 1769--1772. IEEE, 2012.

\bibitem{ijcai2021-441}
Mingkun Xu, Yujie Wu, Lei Deng, Faqiang Liu, Guoqi Li, and Jing Pei.
\newblock Exploiting spiking dynamics with spatial-temporal feature
  normalization in graph learning.
\newblock In Zhi-Hua Zhou, editor, {\em Proceedings of the Thirtieth
  International Joint Conference on Artificial Intelligence, {IJCAI-21}}, pages
  3207--3213. International Joint Conferences on Artificial Intelligence
  Organization, 8 2021.
\newblock Main Track.

\bibitem{lee2020spike}
Chankyu Lee, Adarsh~Kumar Kosta, Alex~Zihao Zhu, Kenneth Chaney, Kostas
  Daniilidis, and Kaushik Roy.
\newblock Spike-flownet: event-based optical flow estimation with
  energy-efficient hybrid neural networks.
\newblock In {\em European Conference on Computer Vision}, pages 366--382.
  Springer, 2020.

\bibitem{horowitz20141}
Mark Horowitz.
\newblock 1.1 computing's energy problem (and what we can do about it).
\newblock In {\em 2014 IEEE International Solid-State Circuits Conference
  Digest of Technical Papers (ISSCC)}, pages 10--14. IEEE, 2014.

\end{thebibliography}

\end{document}